\pdfoutput=1
\documentclass{article}

\usepackage{microtype}
\usepackage{graphicx}
\usepackage{subfigure}
\usepackage{psfrag}
\usepackage{booktabs} %
\usepackage{amsmath}
\usepackage{makecell}
\usepackage{multirow}

\usepackage{hyperref}

\usepackage[accepted]{icml2021}

\begin{document}
\newcommand\supplementalref[1]{in Appendix}  
\newcommand{\omt}[1]{}
\newcommand{\todo}[1]{({\color{red}{ToDo}}: {#1} {\color{red}{***}}}
\newcommand{\bba}{\bar{\bar{\aa}}}
\newcommand{\ba}{\bar{\aa}}
\newcommand{\twostates}[2]{\left(\begin{smallmatrix} #1 \\ #2 \end{smallmatrix}\right)}
\newcommand{\act}[1]{\sigma({#1})}
\newcommand{\numlayers}{n}
\newcommand{\numstates}{k}
\newcommand{\numclasses}{d}
\renewcommand{\vec}[1]{\boldsymbol{#1}}
\newcommand{\pd}[2]{\frac{\partial #1}{\partial #2}}
\newcommand{\tensor}[1]{\hat{\boldsymbol{#1}}}
\newcommand{\sourdough}{BLUR }
\newcommand{\sourdoughlong}{BLUR (Bidirectional Learned Update Rules) }
\newcommand{\longsourdough}{bidirectional learned update rules (BLUR) }

\newcommand{\A}{\ensuremath{\mathbf{A}}}
\newcommand{\B}{\ensuremath{\mathbf{B}}}
\newcommand{\C}{\ensuremath{\mathbf{C}}}
\newcommand{\D}{\ensuremath{\mathbf{D}}}
\newcommand{\E}{\ensuremath{\mathbf{E}}}
\newcommand{\F}{\ensuremath{\mathbf{F}}}
\newcommand{\G}{\ensuremath{\mathbf{G}}}
\newcommand{\HH}{\ensuremath{\mathbf{H}}}
\newcommand{\I}{\ensuremath{\mathbf{I}}}
\newcommand{\J}{\ensuremath{\mathbf{J}}}
\newcommand{\K}{\ensuremath{\mathbf{K}}}
\newcommand{\LL}{\ensuremath{\mathbf{L}}}
\newcommand{\M}{\ensuremath{\mathbf{M}}}
\newcommand{\N}{\ensuremath{\mathbf{N}}}
\newcommand{\OO}{\ensuremath{\mathbf{O}}}
\newcommand{\PP}{\ensuremath{\mathbf{P}}}
\newcommand{\Q}{\ensuremath{\mathbf{Q}}}
\newcommand{\RR}{\ensuremath{\mathbf{R}}}
\renewcommand{\SS}{\ensuremath{\mathbf{S}}}
\newcommand{\T}{\ensuremath{\mathbf{T}}}
\newcommand{\U}{\ensuremath{\mathbf{U}}}
\newcommand{\V}{\ensuremath{\mathbf{V}}}
\newcommand{\W}{\ensuremath{\mathbf{W}}}
\newcommand{\X}{\ensuremath{\mathbf{X}}}
\newcommand{\Y}{\ensuremath{\mathbf{Y}}}
\newcommand{\Z}{\ensuremath{\mathbf{Z}}}
\renewcommand{\aa}{\ensuremath{\mathbf{a}}}
\renewcommand{\b}{\ensuremath{\mathbf{b}}}
\renewcommand{\c}{\ensuremath{\mathbf{c}}}
\newcommand{\dd}{\ensuremath{\mathbf{d}}}
\newcommand{\e}{\ensuremath{\mathbf{e}}}
\newcommand{\f}{\ensuremath{\mathbf{f}}}
\newcommand{\g}{\ensuremath{\mathbf{g}}}
\newcommand{\h}{\ensuremath{\mathbf{h}}}
\newcommand{\bk}{\ensuremath{\mathbf{k}}}
\newcommand{\bl}{\ensuremath{\mathbf{l}}}
\newcommand{\m}{\ensuremath{\mathbf{m}}}
\newcommand{\n}{\ensuremath{\mathbf{n}}}
\newcommand{\p}{\ensuremath{\mathbf{p}}}
\newcommand{\q}{\ensuremath{\mathbf{q}}}
\newcommand{\rr}{\ensuremath{\mathbf{r}}}
\newcommand{\sss}{\ensuremath{\mathbf{s}}}  %
\renewcommand{\t}{\ensuremath{\mathbf{t}}}
\newcommand{\uu}{\ensuremath{\mathbf{u}}}
\newcommand{\vv}{\ensuremath{\mathbf{v}}}
\newcommand{\w}{\ensuremath{\mathbf{w}}}
\newcommand{\x}{\ensuremath{\mathbf{x}}}
\newcommand{\y}{\ensuremath{\mathbf{y}}}
\newcommand{\z}{\ensuremath{\mathbf{z}}}
\newcommand{\0}{\ensuremath{\mathbf{0}}}
\newcommand{\1}{\ensuremath{\mathbf{1}}}

\newcommand{\balpha}{\ensuremath{\boldsymbol{\alpha}}}
\newcommand{\bbeta}{\ensuremath{\boldsymbol{\beta}}}
\newcommand{\bdelta}{\ensuremath{\boldsymbol{\delta}}}
\newcommand{\bepsilon}{\ensuremath{\boldsymbol{\epsilon}}}
\newcommand{\binfty}{\ensuremath{\boldsymbol{\infty}}}
\newcommand{\bkappa}{\ensuremath{\boldsymbol{\kappa}}}
\newcommand{\blambda}{\ensuremath{\boldsymbol{\lambda}}}
\newcommand{\bmu}{\ensuremath{\boldsymbol{\mu}}}
\newcommand{\bnu}{\ensuremath{\boldsymbol{\nu}}}
\newcommand{\bphi}{\ensuremath{\boldsymbol{\phi}}}
\newcommand{\bpi}{\ensuremath{\boldsymbol{\pi}}}
\newcommand{\bpsi}{\ensuremath{\boldsymbol{\psi}}}
\newcommand{\brho}{\ensuremath{\boldsymbol{\rho}}}
\newcommand{\bsigma}{\ensuremath{\boldsymbol{\sigma}}}
\newcommand{\btau}{\ensuremath{\boldsymbol{\tau}}}
\newcommand{\btheta}{\ensuremath{\boldsymbol{\theta}}}
\newcommand{\bxi}{\ensuremath{\boldsymbol{\xi}}}
\newcommand{\bzeta}{\ensuremath{\boldsymbol{\zeta}}}

\newcommand{\bDelta}{\ensuremath{\boldsymbol{\Delta}}}
\newcommand{\bGamma}{\ensuremath{\boldsymbol{\Gamma}}}
\newcommand{\bLambda}{\ensuremath{\boldsymbol{\Lambda}}}
\newcommand{\bPhi}{\ensuremath{\boldsymbol{\Phi}}}
\newcommand{\bPi}{\ensuremath{\boldsymbol{\Pi}}}
\newcommand{\bPsi}{\ensuremath{\boldsymbol{\Psi}}}
\newcommand{\bSigma}{\ensuremath{\boldsymbol{\Sigma}}}
\newcommand{\bTheta}{\ensuremath{\boldsymbol{\Theta}}}
\newcommand{\bXi}{\ensuremath{\boldsymbol{\Xi}}}
\newcommand{\bUpsilon}{\ensuremath{\boldsymbol{\Upsilon}}}

\newcommand{\bbC}{\ensuremath{\mathbb{C}}}
\newcommand{\bbH}{\ensuremath{\mathbb{H}}}
\newcommand{\bbN}{\ensuremath{\mathbb{N}}}
\newcommand{\bbR}{\ensuremath{\mathbb{R}}}
\newcommand{\bbS}{\ensuremath{\mathbb{S}}}
\newcommand{\bbZ}{\ensuremath{\mathbb{Z}}}

\newcommand{\calA}{\ensuremath{\mathcal{A}}}
\newcommand{\calB}{\ensuremath{\mathcal{B}}}
\newcommand{\calC}{\ensuremath{\mathcal{C}}}
\newcommand{\calD}{\ensuremath{\mathcal{D}}}
\newcommand{\calE}{\ensuremath{\mathcal{E}}}
\newcommand{\calF}{\ensuremath{\mathcal{F}}}
\newcommand{\calG}{\ensuremath{\mathcal{G}}}
\newcommand{\calH}{\ensuremath{\mathcal{H}}}
\newcommand{\calI}{\ensuremath{\mathcal{I}}}
\newcommand{\calJ}{\ensuremath{\mathcal{J}}}
\newcommand{\calK}{\ensuremath{\mathcal{K}}}
\newcommand{\calL}{\ensuremath{\mathcal{L}}}
\newcommand{\calM}{\ensuremath{\mathcal{M}}}
\newcommand{\calN}{\ensuremath{\mathcal{N}}}
\newcommand{\calO}{\ensuremath{\mathcal{O}}}
\newcommand{\calP}{\ensuremath{\mathcal{P}}}
\newcommand{\calR}{\ensuremath{\mathcal{R}}}
\newcommand{\calS}{\ensuremath{\mathcal{S}}}
\newcommand{\calT}{\ensuremath{\mathcal{T}}}
\newcommand{\calU}{\ensuremath{\mathcal{U}}}
\newcommand{\calV}{\ensuremath{\mathcal{V}}}
\newcommand{\calW}{\ensuremath{\mathcal{W}}}
\newcommand{\calX}{\ensuremath{\mathcal{X}}}
\newcommand{\calY}{\ensuremath{\mathcal{Y}}}

\newcommand{\abs}[1]{\left\lvert#1\right\rvert}
\newcommand{\norm}[1]{\left\lVert#1\right\rVert}
\newcommand{\ceil}[1]{\lceil#1\rceil}
\newcommand{\floor}[1]{\lfloor#1\rfloor} 

\twocolumn[
\icmltitle{Meta-Learning Bidirectional Update Rules}

\icmlsetsymbol{equal}{*}

\begin{icmlauthorlist}
\icmlauthor{Mark Sandler}{goog}
\icmlauthor{Max Vladymyrov}{goog}
\icmlauthor{Andrey Zhmoginov}{goog}
\icmlauthor{Nolan Miller}{goog}
\icmlauthor{Andrew Jackson}{goog}
\icmlauthor{Tom Madams}{goog}
\icmlauthor{Blaise Ag{\"u}era y~Arcas}{goog}
\end{icmlauthorlist}

\icmlaffiliation{goog}{Google Research}

\icmlcorrespondingauthor{Mark Sandler}{sandler@google.com}

\icmlkeywords{meta-learning, backpropagation, optimization}

\vskip 0.3in
]

\printAffiliationsAndNotice{}  %

\begin{abstract}
In this paper, we introduce a new type of generalized neural network where neurons and synapses maintain multiple states. We show that classical gradient-based backpropagation in neural networks can be seen as a special case of a two-state network where one state is used for activations and another for gradients, with update rules derived from the chain rule. In our generalized framework, networks have neither explicit notion of nor ever receive gradients. The synapses and neurons are updated using a bidirectional Hebb-style update rule parameterized by a shared low-dimensional ``genome''. We show that such genomes can be meta-learned from scratch, using either conventional optimization techniques, or evolutionary strategies, such as CMA-ES. Resulting update rules generalize to unseen tasks and train faster than gradient descent based optimizers for several standard computer vision and synthetic tasks.
\end{abstract}

\section{Introduction}
Neural networks revolutionized the way ML systems are built today.  Advances in neural design patterns, training techniques, and hardware performance allowed ML to solve tasks that seemed
hopelessly out of reach less than ten years ago. However, despite the rapid progress, their
basic neuron-synapse design has remained fundamentally unchanged for nearly six decades, since
the introduction of perceptron models in the 50s and 60s~\citep{minsky69perceptrons,rosenblatt1957perceptron} that modeled the
complex biology of a synapse firing as a simple combination of a weight and a bias combined with a  non-linear activation function. 

With such models, the next question was ``How should we best find the optimal weights and biases?'' and great successes have come from the use of stochastic gradient descent, traced back to \citet{SGD-robbins1951}. Since its introduction, many of the more recent remarkable improvements can be attributed to improving the efficiency of the gradient signal: adjusting connectivity patterns such as in convolutional neural networks and residual layers~\cite{ResNet}, improved optimizer design \cite{Kingma2014AdamAM,adagrad,schmidt2020descending}, and normalization methods such as \cite{batchnorm, instancenorm}. All these methods improve the learning characteristics of the networks and enable scaling to larger networks and more complex problems. 

However the underlying principle behind these methods remained the same: minimize an engineered loss function using gradient descent. Instead, we propose a different approach. While we still follow the general strategy of forwards and backwards signal transmission, we learn the rules governing both forward and back-propagation of neuron activation from scratch. The key enabling factor here is a generalization where each neuron can have multiple states. 

We define a \emph{space} of possible transformations that specify the interaction between neurons' feed-forward and feedback signals. The matrices controlling these interactions are meta-parameters that are shared across both layers and tasks. We term these meta-parameters a ``genome''. This reframing opens up a new, more generalized space of neural networks, allowing the introduction of arbitrary numbers of states and channels into neurons and synapses, which have their analogues in biological systems, such as the multiple types of neurotransmitters, or chemical vs.\ electrical synapse transmission.

Our framework, which we call \sourdoughlong describes a general set of multi-state update rules that are capable to train networks to learn new tasks without ever having access to explicit gradients. We demonstrate that through meta-learning \sourdough can learn effective genomes with just a few training tasks. Such genomes can be learned using off-the-shelf optimizers or evolutionary strategies. We show that such genomes can train networks on unseen tasks faster than comparably sized gradient networks. The learned genomes can also generalize to architectures unseen during the meta-training.

\omt{
tl;dr for reviewers

Mention: 
- no gradients are needed, no loss function is needed
- gradient descent is a special case of two-state system.
- we are inspired by chain rule and attempt to generalize it,
- gradient descent as special case 
- meta-learning is a go-to algorithm
- we meta-learning the hyperparameters of the network as well

AJ: do we mention the large number of pre-tuned optimizers that may resemble points in our 'genome space' ?
AJ: We should call out the importance of the normalization/saturation and why it keeps with the 'biologically inspired'
}

\section{Related Work}
\paragraph{Alternatives to stochastic gradient descent}

Replacing the backpropagation of loss function gradients with a different signal has been explored before.
One example is a family of methods building upon {\em difference target propagation} \cite{bengio2014autoencoders,lee2015difference}, which uses the desired output ``targets'' or ``errors'' as the primary signals communicated during the backpropagation stage.
A recent paper by \citet{ahmad2020gait} builds on the target propagation approach and proposes an alternative, a more biologically plausible {\em local update} that is shown to be equivalent to the SGD update.
Similarly, SGD can be expressed as a special case of a more general family of update rules, which we explore for the best-performing learning algorithm.

Another related line of work builds upon the {\em feedback alignment} method that replaces backwards weights in SGD with fixed random matrices \cite{lillicrap2016random,nokland2016direct}.
In \citet{akrout2019deep}, the authors extend this idea by separately evolving backward weights for automatic ``synchronization'' of forward and backward weights. In addition, \citet{xiao2018biologically} have demonstrated empirically that many of the biologically inspired methods listed above train {\em at most} as good as SGD.

The bidirectional nature of networks was explored in \cite{pontes2019bidirectional,adigun2019bidirectional}, where the backward  pass is treated as a generative inference rather than learning update.

The common feature of approaches above is that all these methods rely on a standard forward pass, and a hand-designed backward pass.
While some variants of feedback alignment are contained within the family of update rules we explore in this paper, in contrast to them our approach learns the meta-parameters that control both the forward and backward passes.

Another line of inquiry are methods that utilize greedy layer-wise training \cite{bengio2006greedy,belilovsky2019greedy,lowe2019putting,xiong2020loco}, a pseudoinverse \cite{ranganathan2020zorb}, or like \citet{taylor2016training} reformulate the end-to-end training as a constrained optimization problem with additional variables.
While these methods do not share as many common elements with the present approach, we believe that exploring layer-wise training in conjunction with our method is a promising direction for future research.

\vspace{-2ex}
\paragraph{Meta-Learning}
Recently, researchers have turned to meta-learning \cite{schmidhuber1987evolutionary} approaches that aim on improving existing functional modules \cite{munkhdalai2019metalearned} or learning methods by meta-training optimal hyper-parameters (often called meta-parameters) in a problem-independent way \cite{andrychowicz2016learning,wichrowska2017learned,maheswaranathan2020reverse,metz2019understanding,metz2020tasks}. The important difference with our method is that these work by directly using the gradient of a given loss function, whereas our method proposes a holistic learning framework that does not correspond to any known optimizer or even rely on a predefined loss function.

It has long been recognized that SGD is a biologically implausible mechanism  \cite{bengio2015towards}.
Another direction that is explored in the literature is more biologically plausible mechanisms \cite{soltoggio2018born} including those based on {\em Hebb's rule} \cite{hebb1949organization} and its modification {\em Oja's rule} \cite{oja1982simplified}. 
Meta-learning has been used to learn the plasticity of similar update rules \cite{miconi2018differentiable,miconi2019backpropamine,lindsey2020learning,confavreux2020approach,najarro2020meta} as well as different neuromodulation mechanisms \cite{bengio1995search,norouzzadeh2016neuromodulation,velez2017diffusion,wilson2018neuromodulated}.
Recent work by \citet{camp2020continual} followed a related direction learning the sub-structure of individual neurons by representing them as perceptron networks while keeping gradient-based backpropagation.
In yet another alternative approach, \citet{kaplanis2018continual} proposed to introduce more nuanced memory to synapses, while in \citet{randazzo2020mplp}, authors meta-learn parameters of a complex message-passing learning algorithm that replaces the backpropagation while leaving the forward pass intact.

\citet{kirsch2020meta} propose a generalized learning algorithm based on a set of RNNs that, similar to our framework, does not use any gradients or explicit loss function, yet is able to approximate forward pass and backpropagation solely from forward activations of RNNs.
Our system, in contrast, does not use RNNs and explicitly leaves (meta-parametrized) bidirectional update rules in place. 

Different from traditional meta-learning, \citet{real2020automl} and \citet{ha2016hypernetworks} devise a specialized learning algorithm to directly find the optimal parameters of another target algorithm that they want to learn.

To the best of our knowledge, our paper is the first work that customizes {\em both} inference and learning passes by successfully finding the update rule for both forward and backward passes that does not rely on neither explicit gradients or a predefined loss function.

\begin{figure*}[t]
\centering
  \includegraphics[width=0.52\textwidth]{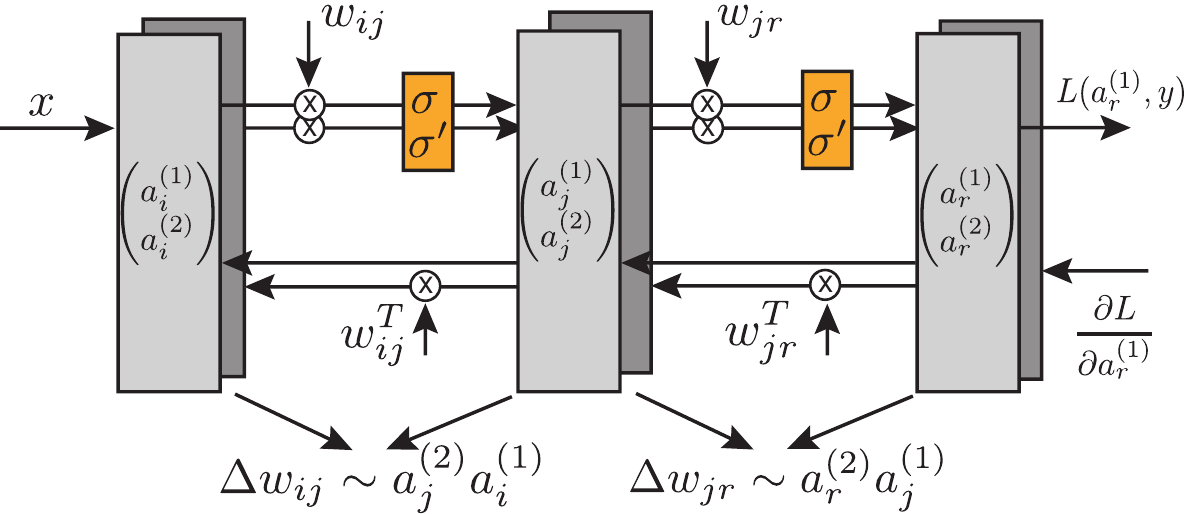} 
  \includegraphics[width=0.47\textwidth]{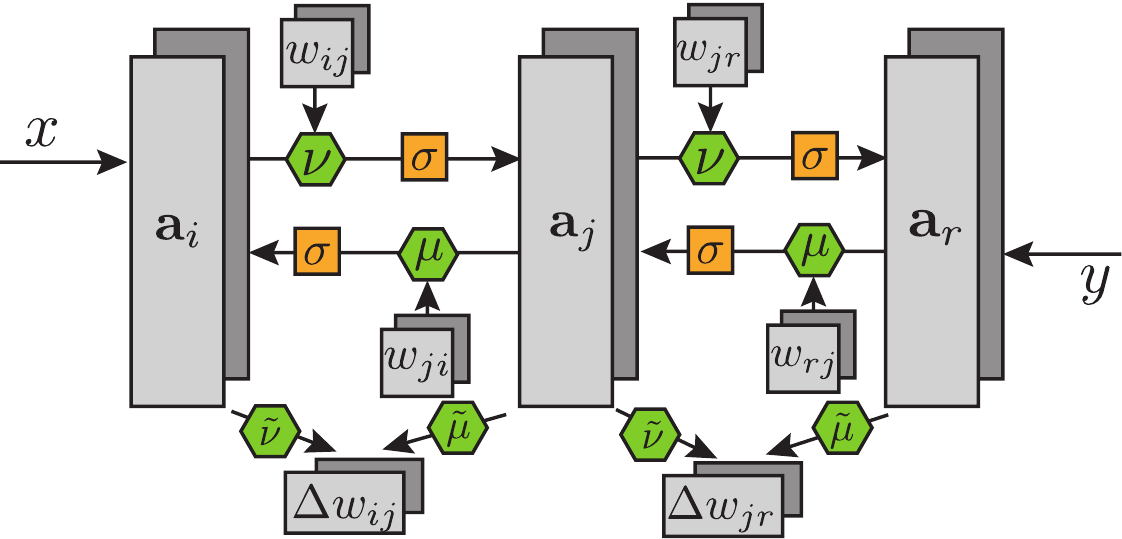} 
  \caption{Generalization of a three layers feed-forward neural networks as a multi-state systems. \emph{Left:} Forward pass and chain-rule backpropagation organized as a generalized two-state network. Arrows indicate the flow of information from forward and backward passes to synapse updates. \emph{Right:} Our proposed generalized formulation. Green nodes, defined in the genome, control the amount of mixing between the states. They are fixed during the synapse update (inner-loop) and are optimized during the meta-training (outer-loop). Grey boxes indicate multi-state variables. Orange boxes represent activation functions. Notice the symmetry between forward and backward passes.}
  \label{f:backprop_vs_sourdough} 
\end{figure*}

\def\numneurons{n}
\def\numfeatures{l}
\def\Genome{{\cal G}}
\def\Ex{\mathbf{E}}
\def\numunrolls{s}
\newcommand{\mc}[1]{\mathcal{#1}}

\section{Learning a New Type of Neural Network}
\subsection{A generalization of gradient descent using neuronal state}
\label{sec:backprop_genome}
To learn a new type of neural network we need to formally define the space of possible configurations. Our proposed space is a generalization of classical artificial neural networks, with inspiration drawn from biology.  For the purpose of clarity, in this section we modify the notation by abstracting from the standard layer structure of a neural network, and instead assume our network is essentially a \emph{bag-of-neurons} $\mc{N}$ of  $\numneurons$ neurons with a connectivity structure defined by two functions: ``upstream'' neurons $I(i)\subset \mc{N}$ that send their outputs to $i$, and the set of ``downstream'' neurons $J(i)\subset \mc{N}$ that receive the output of $i$ as one of their inputs. Thus the synapse weight matrix $w_{ij}$ can encode separate weights for forward and backward connections. 

Normally we think of a neuron in an artificial neural network as having a single scalar value, it turns out that one state is not enough once we incorporate a back-propagation signal, which uses both feed-forward state and feedback state to propagate through the network.  
The standard forward pass over a densely connected neural network updates the state of each neuron $j\in \mc{N}$ according to
\begin{equation}
    \label{eq:ff}
    \textstyle{h_j = \mathop{\sigma}\left( \sum_{i\in I(j)} w_{ij} h_i \right),}
\end{equation}

where $h_j$ is the activation for neuron $j\in \mc{N}$ resulting from applying a function $\sigma(\cdot)$ to the product of the network weights $w_{ij}$ and the incoming stimulus from $I(j)$. Here and further we combine the bias and the weights into a single vector by adding a unit element at the end of each hidden vector. For the first layer, the activation is given by the input batch. Let us define $h'_j := \mathop{\sigma'}\left( \sum_{i\in I(j)} w_{ij} h_{i} \right)$ as a derivative of the activation with respect to its argument. Then we can write chain rule, describing the back-propagation of a loss function as:
\begin{equation}
    \label{eq:bp}
    \textstyle{\frac{\partial{L}}{\partial h_i} = \sum_{j\in J(i)} w_{ij} \frac{\partial{L} }{\partial h_j} h'_j.}
\end{equation}
For the last layer, the first step of backpropagation is given by the derivative of the loss function with respect to the last activations. After the backpropagation, using the notation above, the gradient descent updates the synapses $w_{ij}$ using
\begin{equation}
    \label{eq:weight_update}
    \textstyle{w_{ij} \leftarrow w_{ij} - \tilde\eta\frac{\partial{L}}{\partial h_j} h'_j h_i, }
\end{equation}
where $\tilde\eta$ is a learning rate. 

Notice that the update has a form of the Hebbian learning rule \citep{hebb1949organization} with pre-synaptic activation given by $h_i$ and post-synaptic one given by $\frac{\partial{L}}{\partial h_j} h'_j$. We can make this connection even more explicit using neurons with two states, i.e., activations above are now replaced with a two-dimensional vector $\vec{a}_i = (a_i^{(1)}, a_i^{(2)})$.
One of these states would be used for a feed-forward signal and another for a back-propagated feedback signal. During the forward pass we set $\vec{a}_j \leftarrow (h_j, h'_j)$ and during the backward pass the second state is updated multiplicatively using \eqref{eq:bp} as
$a_i^{(2)} \leftarrow \frac{\partial{L}}{\partial h_i} h'_i = a_i^{(2)}\sum_{j \in J(i)} w_{ij} a_j^{(2)}$.
Then the synapse update is given by $w_{ij} \leftarrow w_{ij} - \tilde\eta a^{(2)}_ja^{(1)}_i$. The left side of Figure \ref{f:backprop_vs_sourdough} demonstrates the described operations as a two-state neural network.

We can further generalize the learning procedure with the following constant matrices: $\nu=\left(\begin{smallmatrix} 1 & 0 \\ 1 & 0 \end{smallmatrix} \right)$, $\mu = \twostates{0}{1}$, $\tilde\nu = \twostates{1}{0}$, $\tilde\mu = \twostates{0}{1}$ and a generalized binary activation function $\phi\big(\twostates{x}{y}\big)=\twostates{\sigma(x)}{\sigma'(y)}$. Then the operations above can be equivalently rewritten as
\vspace{-3ex}

\begin{equation}
    \label{eq:bp-two-states}
    \begin{array}{llcl}
    \hspace{-2ex} \text{\small Forward pass:} & a^c_j &\leftarrow& \phi^c\big(\sum\limits_{i\in I(j),d}w_{ij}a^d_i\nu^{cd}\big)\\
    \hspace{-2ex} \text{\small Backward pass:} & a^{(2)}_i &\leftarrow& a^{(2)}_i\sum\limits_{j\in J(i),d}w_{ij}a^{d}_j\mu^d \\ 
    \hspace{-2ex} \text{\small Weights update:} & w_{ij} &\leftarrow& w_{ij} - \tilde\eta \sum\limits_{c,d} a^{c}_j\tilde{\mu}^ca^{d}_i\tilde\nu^d. \\
    \end{array}
\end{equation}

Here $c, d = \{1,2\}$ represent the states of the network and we use superscript to index over the states. 
Thus, traditional gradient backpropagation can be expressed as a general two-state network, whose update rules are controlled by a predefined set of very low-dimensional matrices $\{\nu, \mu, \tilde\nu, \tilde\mu, \tilde\eta\}$. These matrices are fixed during the weight update phase and optimized during the meta-optimization to achieve a more general update rules.

\subsection{Multi-state bidirectional update rules}

The two-state interpretation of the backpropagation algorithm outlined above is asymmetrical and contains several potentially biologically implausible design details like the use of the same weight matrix on the forward and backward passes and a multiplicative update during the backpropagation phase.
Being inspired by this update mechanism, we propose a general family of \longsourdough that: (a) use multi-channel asymmetrical synapses, (b) use the same update mechanisms on the forward and backward paths, and finally (c) allow for information mixing between different channels of each neuron.
In its final state, this family can be described by the following equations:
\begin{alignat}{3}
\label{eq:sourdough}
    & \text{\small Forward pass:} \quad && a^c_j & \leftarrow & \sigma\big(f a^c_{j} + \eta \sum\limits_{i \in I(j),d} w^{c}_{ij} \nu^{cd} a^d_{i}\big)\\
    \label{eq:sourdough-back}
    & \text{\small Backward pass:} \quad && a^c_i & \leftarrow & \sigma\big(f a^c_{i} + \eta \sum\limits_{j \in J(i),d} w^c_{ji}  \mu^{cd} a^d_{j}\big) \\ 
    & \text{\small Weights update:} \quad && w^c_{ij} & \leftarrow & \tilde{f}w^c_{ij} + \tilde{\eta} \sum\limits_{e, d} a^e_{i} \label{eq:sourdough-last}\tilde{\nu}^{ec} \cdot \tilde{\mu}^{cd} a^d_{j}.
\end{alignat}    

The right side of Fig.~\ref{f:backprop_vs_sourdough} demonstrates the proposed framework and Table~\ref{tab:vars} tracks the description and dimensionality of variables used in the formulae above.  The matrices $\{f, \tilde f, \nu, \tilde\nu, \mu, \tilde\mu, \eta, \tilde\eta\}$ used in (\ref{eq:sourdough}--\ref{eq:sourdough-last}) form our complete \emph{genome} $\Genome$.

\begin{table}[t]
    \centering
     \begin{tabular}{c|c|p{3.5cm}}
     Name & Dimension & Description \\\hline
     \multicolumn{3}{c}{Constants} \\\hline
     $\numneurons$ &-& total number of neurons. \\
     $k$ &-& total number of states. \\\hline
     \multicolumn{3}{c}{Network params} \\\hline
     $a_i^c$ & $i\in [n], c\in [k]$ & state $c$ of neuron $i$. \\     
     $w^c_{ij}$ & $i, j\in [\numneurons], c\in [k]$ & channel $c$ of synapse between $i$ and $j$. \\\hline
     \multicolumn{3}{c}{Meta-learning params (genome)} \\\hline
     $f$, $\eta$ & $1$ & neuron \texttt{\small forget} and \texttt{\small update} gates. \\
     $\tilde{f}$, $\tilde\eta$ & $1$ & synapses \texttt{\small forget} and \texttt{\small update} gate. \\
     $\nu^{cd}, \mu^{cd}$ & $c\in [k], d\in [k]$ & forward/backward neuron transform matrix. \\
     $\tilde\nu^{cd}, \tilde\mu^{cd}$ & $c\in [k], d\in [k]$ & pre-\ and post-synaptic transform matrix.
    \end{tabular}
    \caption{Description and dimensions of variables.}
    \label{tab:vars}
\end{table}

Here we make the following generalizations with respect to the backpropagation update rules:
\begin{itemize}
    \item The neuron transform matrices $\nu, \mu$ and synapse transform matrices $\tilde \nu, \tilde \mu$ all have dimension $k \times k$ and allow for mixing of every input state to every output state as well as possibility using more than two states in the genome.
    \item We expand the genome to include $f, \eta, \tilde f, \tilde \eta$ that control how much of the information is \emph{forgotten} and how much is being \emph{updated} after each step. A similar approach has been studied in \citet{ravi2017optimization}, however we learn these scalars directly and do not model them as a function of a previous iteration.
    \item We propose an additive update for both neurons and synapses. Note that in order to generalize to backpropagation, an additive update for the backward pass has to be replaced with a multiplicative one and applied only to the second state. Experimentally, we discovered that both additive and multiplicative updates perform similarly.
    \item We extend the activation function to be applied on both forward and backward pass and, to make things simple, make it unitary (same function applied to every state).
    \item We generalize the synapse matrices to be asymmetric for a forward and backward pass ($w_{ij}\neq w_{ji}$) as well as contain more than one state. Symmetric weight matrices are ordinarily used for deep learning, but distinct weight matrices are more biologically plausible. 
    \item The synapse update has a general form of a Hebbian-update rule mixing pre- and post-synaptic activity according to the synapse transform matrices $\tilde \nu, \tilde \mu$.
\end{itemize}

In addition to generalizing existing gradient learning, not relying on gradients in backpropagation has additional benefits. For example, the network doesn't need to have an explicit notion of a final loss function. The feedback (ground truth) can be fed directly into the last layer (e.g.\ by an update to the second state or simply by replacing the second state altogether) and the backward pass would take care of backpropagating it through the layers.

Notice that the genome is defined at the level of individual neurons and synapses and is independent from the network architecture. Thus, the same genome can be trained for different architectures and, more generally, genome trained on one architecture can be applied to a genome with different architectures. We show some examples of this in the experimental section. 

Since the proposed framework can use more than two states, we hypothesize that just as the number of layers relates to the complexity of learning required for an individual task (inner loop of the meta-learning), the number of states might be related to complexity of learning behaviour across the task (outer loop). More informally: synapses regulate ``how hard is a given task'' vs genome's ``how hard it is to learn the task given a variety of other tasks available''.

Our resulting genome now completely describes the communication between individual neurons.  The neurons themselves can be arranged in any of the familiar ways -- in convolutional layers, residual blocks, etc. For the rest of the paper we focus on the simplest types of networks consisting of one or more fully connected layers. 

\subsection{Meta-learning the genome}
Once we have defined the space of possible update rules, the next step is to design an algorithm to find a useful genome capable of successful training and generalization. In this work we concentrate on meta-learning genomes that can solve classification problems with multiple hidden layers. %

For a $\numclasses$-class classification problem with $\numfeatures$-dimensional input, we use the first layer as input and the last layer with $\numclasses$ neurons as predictors. We denote those neurons as $x_1\dots x_\numfeatures$ and  $y_1\dots y_\numclasses$ respectively.

During the learning process, in the forward pass we apply equation \eqref{eq:sourdough} to compute a logit prediction for a given class $i$ to the {\em first} state of the last layer $a^{(1)}_{y_i}$. During the backward pass, we set the {\it second} state $a^{(2)}_{y_i}$ as $1$ or $-1$ from the ground truth based on the class attribute. In experiments with more than two states per neuron, we fill the other states of the last layer with zeros.  We then use equations \eqref{eq:sourdough-back} and \eqref{eq:sourdough-last} to compute updated synapse state.

To evaluate the quality of a genome we apply equations (\ref{eq:sourdough}--\ref{eq:sourdough-last}) for
multiple unroll steps and then test learned synapses on a previously unseen
set of inputs. To meta-learn using SGD we use standard softmax-cross entropy loss: 
$L_{\text{meta}}(\Genome) = \Ex_s  \left[p_i(s)\log a^{(0)}_{y_i}(s)\right]$ where $p_i(s)$ is one-hot vector
representing the true category for a sample $s$, and $a^{(0)}_{y_i}(s)$ is the prediction of that sample from the forward pass, after applying our forward and backward updates for a given number of unroll steps. We then can minimize this function with standard off-the-shelf optimizers to find an optimal genome. 

In section \ref{sec:cma} we also perform experiments using  CMA-ES \cite{cma-es} where we use training accuracy as a fitness metric.

\subsection{Activation normalization and synapse saturation}
    Synapse updates that rely on Hebb's rule alone ($\Delta w_{ij} \sim a_i a_j$) are generally unstable, as network weights $w$ grow monotonically with the training steps.
One way to alleviate this issue while also reducing the sensitivity of network outputs to small synapse perturbations is to use {\em activation normalization}.
Normalization techniques are known to be used by certain biological systems \cite{carandini2012normalization} to calibrate neuron activation to the optimal firing regime and are widely used in conventional deep neural architectures \cite{batchnorm,instancenorm}.
In most of our experiments, we used per-channel normalization similar to batch-normalization to normalize a pre-nonlinearity activation distribution into one with a learnable mean and deviation.
To maintain the symmetry between the forward and backward pass we apply normalization for both forward and backward activations.
This not only helps in training deeper models, but also allows learned update rules to generalize to different input sizes and number of classes.

However, activation normalization alone does not always prevent an unbounded growth of synapse weights, and so another mechanism for weight saturation is necessary.
One such approach is based on using Oja's update rule \cite{oja1982simplified} that modifies Hebb's rule with an additional component that by itself leads to the decay of the singular components of the weight matrix.
One of the most commonly used forms of Oja's update rule is $\Delta w_{ij} = \gamma a_i a_j - \gamma a_j^2 w_{ij}$ however there also exist other forms with similar properties.
In linear systems, the interplay of the excitatory and inhibitory driving forces leads to synapse saturation \cite{oja1982simplified}.
But in our case, the linear component of the update rule $\tilde{f}w$ along with the usage of nonlinearities (like $\tanh$) or activation normalization may prevent Oja's original rule from saturating model weights.
In Appendix~\ref{sec:oja_ext}, we show that the same principles that were used to derive Oja's rule can also be applied to our system and result in the following inhibitory Oja-like update:
\begin{multline}
  \label{eq:new_oja}
  (\Delta w_{ij}^{c})^{\rm Oja} = - (\tilde{f} - 1) w_{ij}^c \sum_{r} (w_{rj}^c)^2 - \\ -
    \tilde{\eta} w_{ij}^c \sum_{r} w_{rj}^c \sum_{e,d} a_{r}^{e} \tilde{\nu}^{ec} \cdot \tilde{\mu}^{cd} a_{j}^{d}.
\end{multline}
The first component of this update usually dominates the second component, so in our experiments we only used this term with an additional learnable multiplier.

Oja's term is generally derived as an inhibitory additive component that acts as a ``counterweight'' to the Hebbian synapse update and keeps the weight norm fixed.
Instead of using such inhibitory terms, we could instead apply normalization and saturating nonlinearities (like $\alpha \tanh (x/\alpha)$ with a learnable $\alpha$ coefficient) directly to the synapses.
In our experiments, we empirically validated that both approaches lead to very similar results and could thus potentially be used interchangeably.

\subsection{Is this update rule still a gradient descent?}
    \begin{figure}
    \centering
    \includegraphics[width=0.3\textwidth]{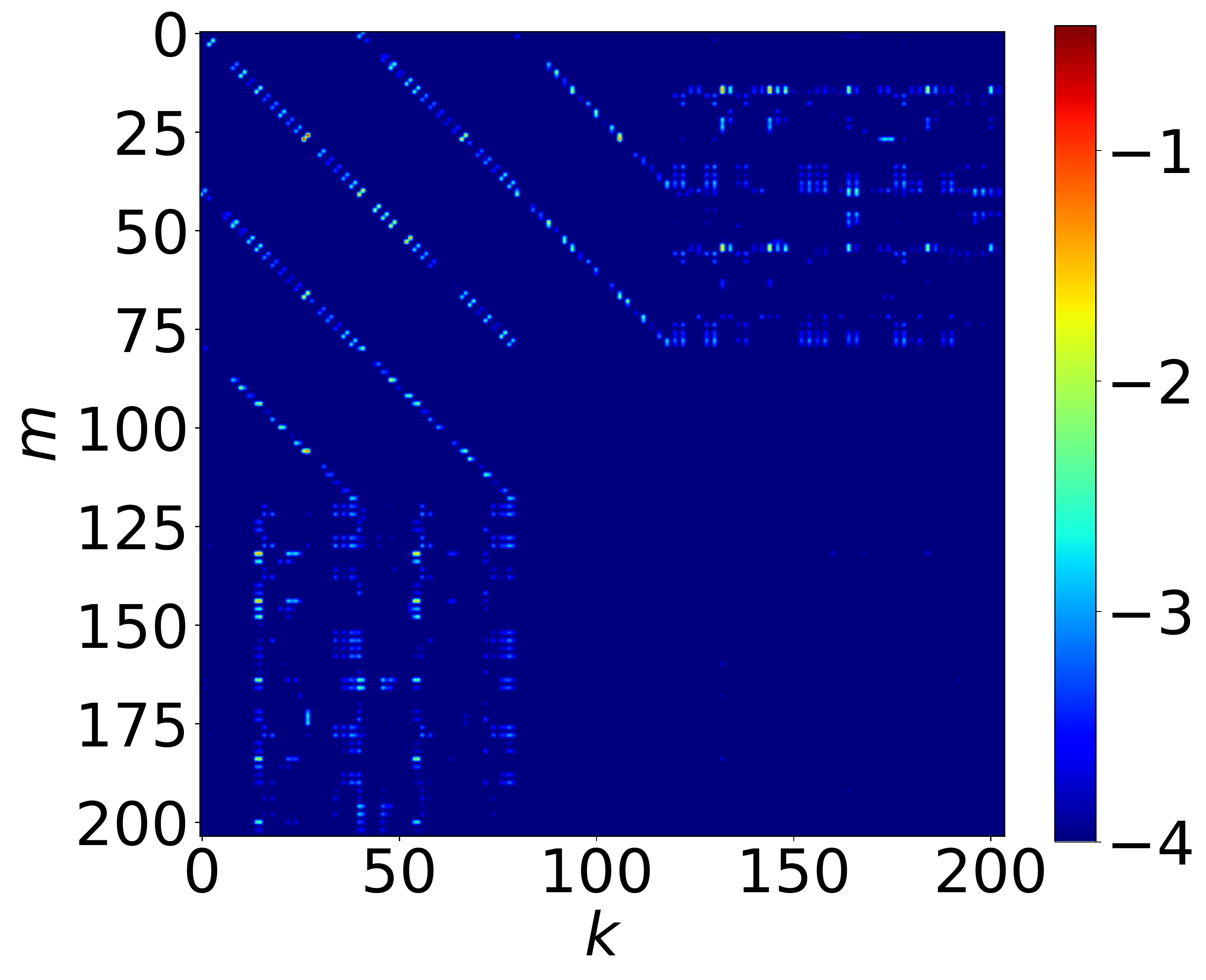}
    \caption{
        Natural logarithm of $|\partial \Delta w_{k} / \partial w_{m}-\partial \Delta w_{m} / \partial w_{k}|$ computed numerically for flattened and concatenated forward weight matrices in a two-state model learning a 2-input Boolean task (e.g. {\tt xor}).
        The model has a single hidden layer of size $20$ and thus $204$ total forward weights and biases.
        The magnitude of the numerical error is estimated to be below $e^{-4}$.
        Bands in the left-upper corner correspond to asymmetries between first-layer connections to the same hidden neurons; other structures appear to arise because of the asymmetries between the first- and the second-layer weights.
    }
    \label{fig:sym_derivatives}
\end{figure}

\begin{figure*}[t]
    \centering
    \begin{tabular}[c]{@{\hspace{-0.005\linewidth}}c@{\hspace{0.01\linewidth}}|c@{\hspace{0.0\linewidth}}}
    Training datasets & Validation datasets \\
        \includegraphics[width=0.65\textwidth]{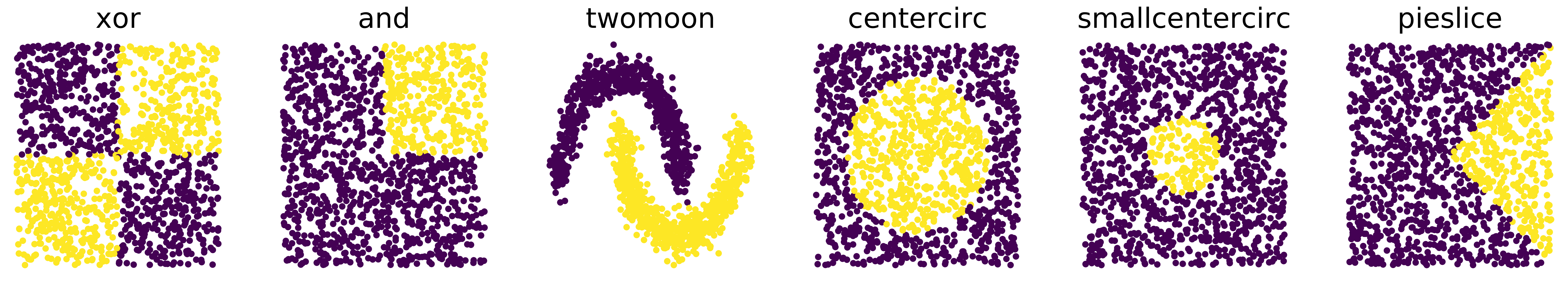} & 
        \includegraphics[width=0.32\textwidth]{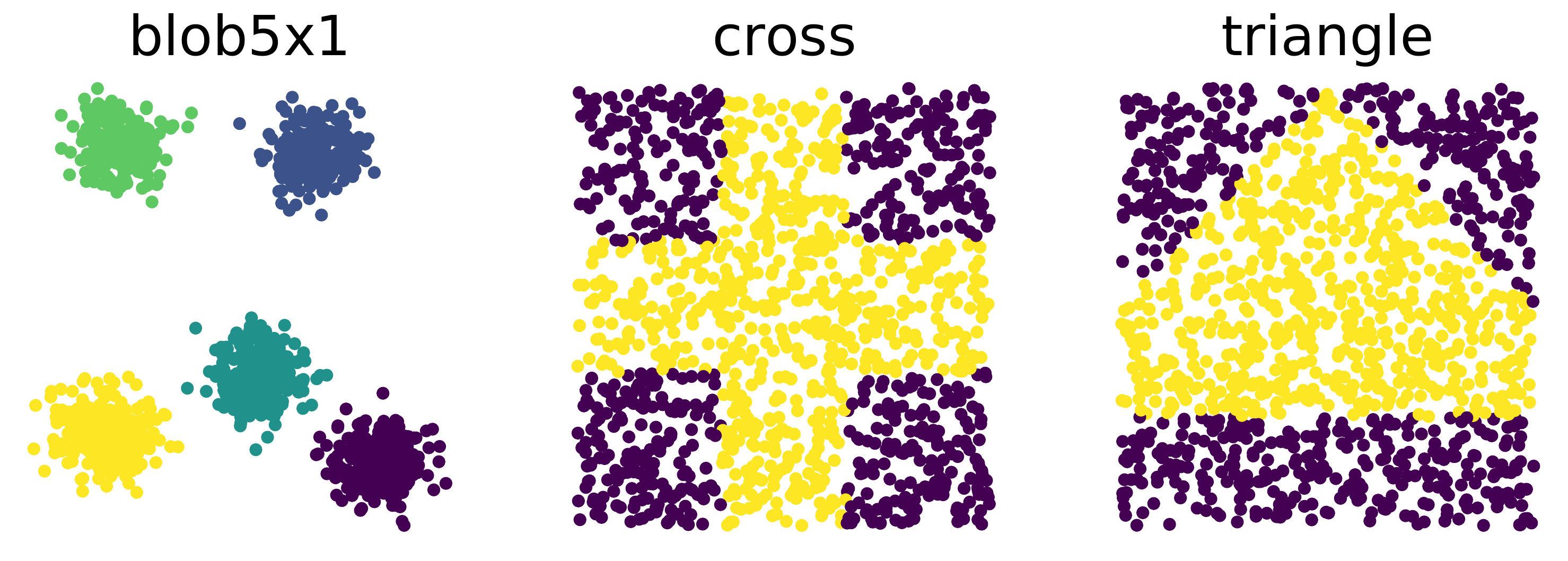} \\
        \includegraphics[width=0.65\textwidth]{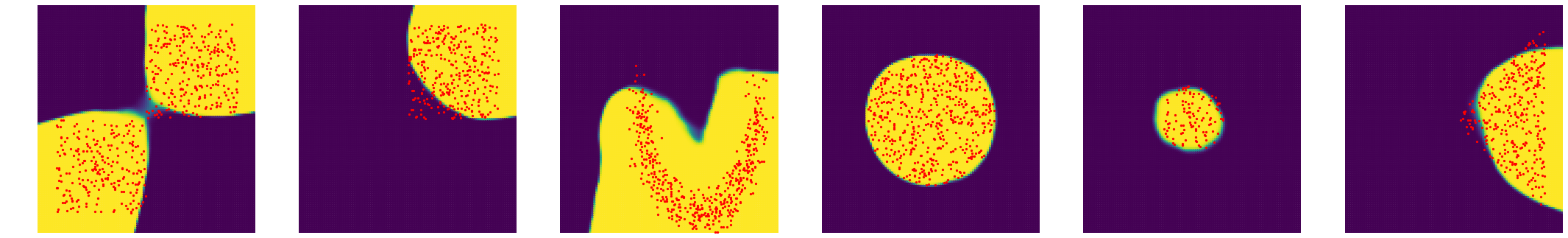} &
        \includegraphics[width=0.32\textwidth]{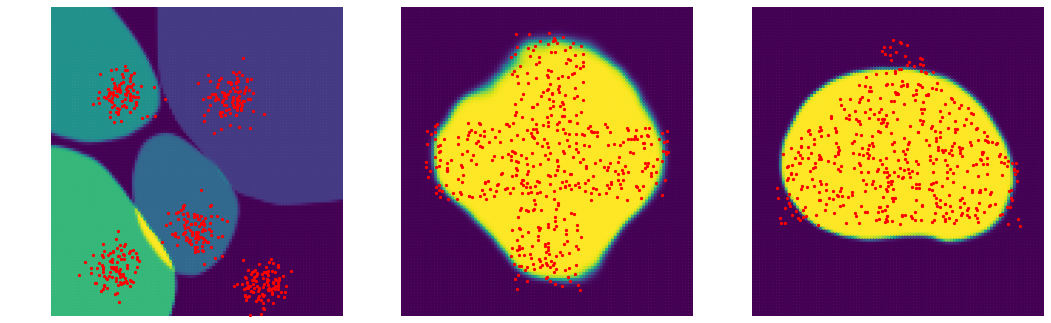}
    \end{tabular}
    \caption{Toy datasets. \emph{Left side:} meta-training datasets, \emph{right side:} meta-validation datasets. \emph{Top row:} training data, \emph{bottom row:} prediction of the trained genome produces on a dense and enlarged grid. }
    \label{boolean-datasets}
\end{figure*}

Once we have used meta-learning to identify a promising genome, one might ask if the resulting training algorithm is in fact identical to a conventional gradient descent with some unknown loss function $L_{\rm equiv}$.
In this section we empirically demonstrate that the answer is generally no.
Consider a full-batch training scenario and let $\Delta w_{ij}^c(\tensor{w};\vec{a})$ be the weight update rule defined by our learned genome. Equivalence to the gradient descent would then mean that
\begin{gather}
    \label{eq:l_gd}
    \Delta w_{ij}^c = -\gamma \pd{L_{\rm equiv}}{w_{ij}^c}
\end{gather}

and therefore
\begin{gather*}
    \pd{\Delta w_{ij}^c}{w_{mn}^d} = -\gamma \frac{\partial^2 L_{\rm equiv}}{\partial w_{ij}^c \partial w_{mn}^d}.
\end{gather*}

Since the partial derivatives are symmetric, we see that
\begin{gather}
    \label{eq:sgd_cond}
    \pd{\Delta w_{ij}^c}{w_{mn}^d} = \pd{\Delta w_{mn}^d}{w_{ij}^c}
\end{gather}

is a necessary condition for the existence of the loss $L_{\rm equiv}$ satisfying Eq.~\eqref{eq:l_gd}.
This condition can be tedious to verify analytically, but we can instead check it numerically.
Computing $|\partial \Delta w_{ij}^c / \partial w_{mn}^d-\partial \Delta w_{mn}^d / \partial w_{ij}^c|$ in a simple experiment with Boolean functions and a single hidden layer of size $20$, we verified that the discovered update rules do not satisfy condition~\eqref{eq:sgd_cond} (see Fig.~\ref{fig:sym_derivatives}) and therefore $L_{\rm equiv}$ does not generally exist for our update rule family.

The observation that learning trajectories obtained with our update rule cannot be recovered using conventional gradient descent does not rule out a potential equivalence to other learning algorithms such as, e.g., a gradient descent with a Riemannian metric $\tensor{g}(\vec{w})$ defined via $\Delta w_i = -\gamma \sum_j g_{ij} \cdot (\partial L_{\rm equiv}/\partial w_j)$
(for details see Appendix~\ref{sec:as_sgd_ext}).
It is also worth noticing that the question of existence of a loss function that is monotonically non-increasing along the training trajectories is directly related to the well-established theory of Lyapunov functions that currently includes many general existence theorems \cite{conley1978isolated,conley1988gradient,farber2003smooth,franks2017notes} and is subject of future work.

    \label{sec:connection-to-gradient-descent}

\section{Experiments}
    \label{sec:experiments}
    \begin{figure*}
    \centering
    \begin{tabular}[c]{@{\hspace{-0.005\linewidth}}c@{\hspace{0\linewidth}}|@{\hspace{0\linewidth}}c@{\hspace{0.0\linewidth}}|@{\hspace{0\linewidth}}c@{\hspace{-0.00\linewidth}}}
    \multicolumn{3}{l}{\hspace{-1.5ex}\includegraphics[width=1.0\textwidth]{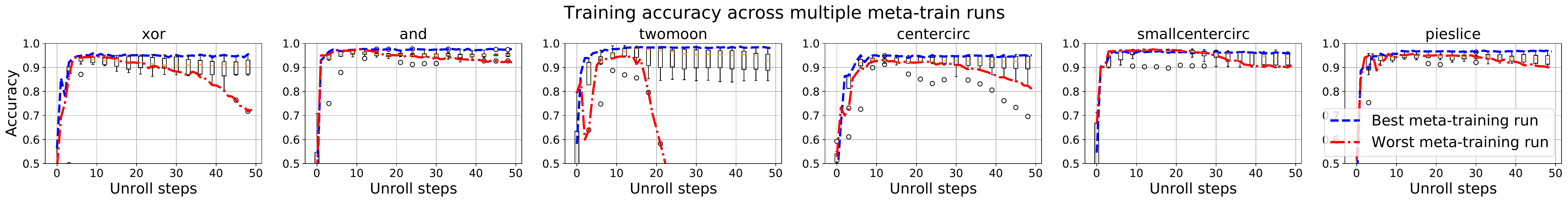}} \\\hline %
    \includegraphics[width=0.41\textwidth]{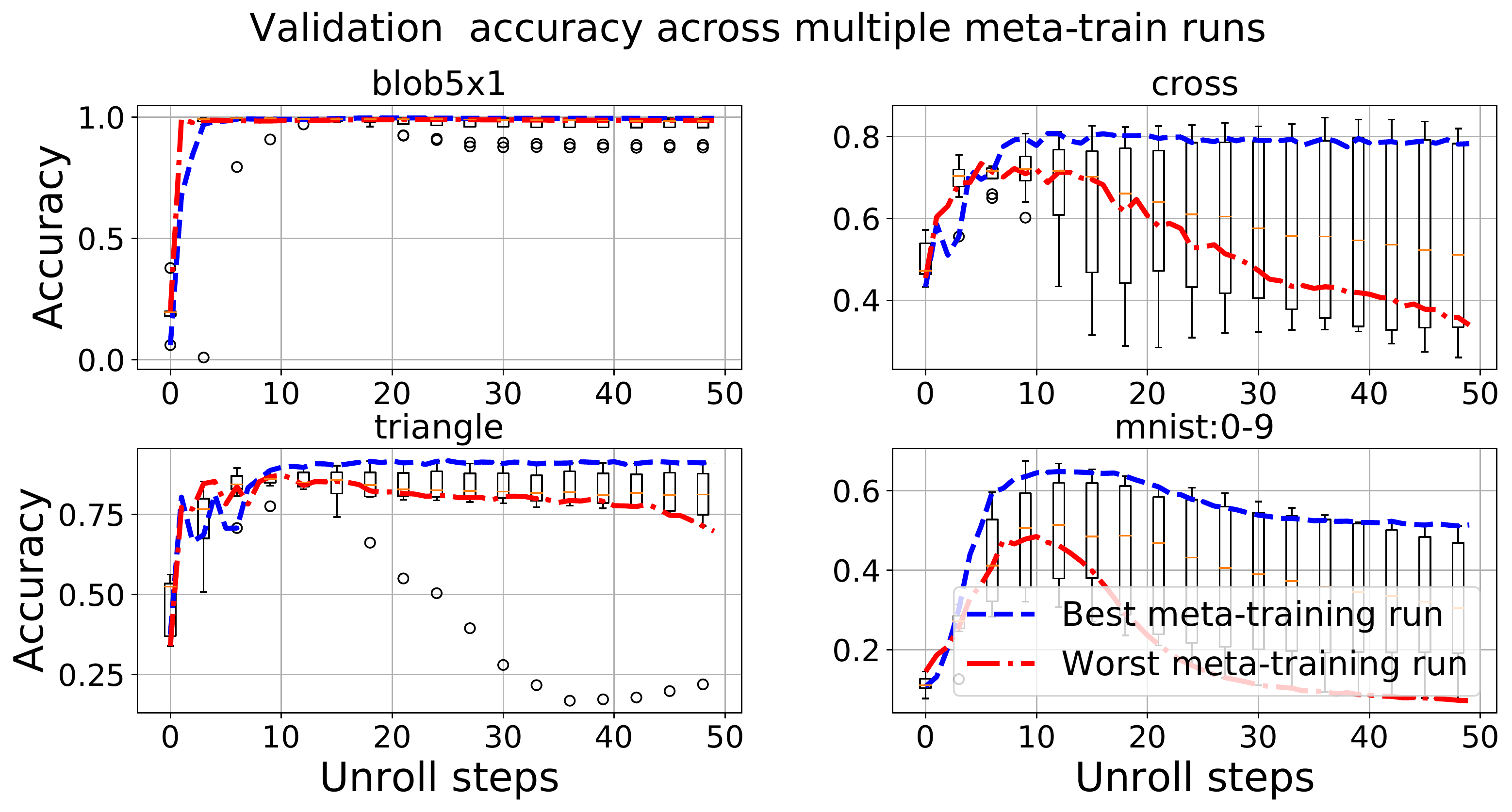} & %
    \includegraphics[width=0.18\textwidth]{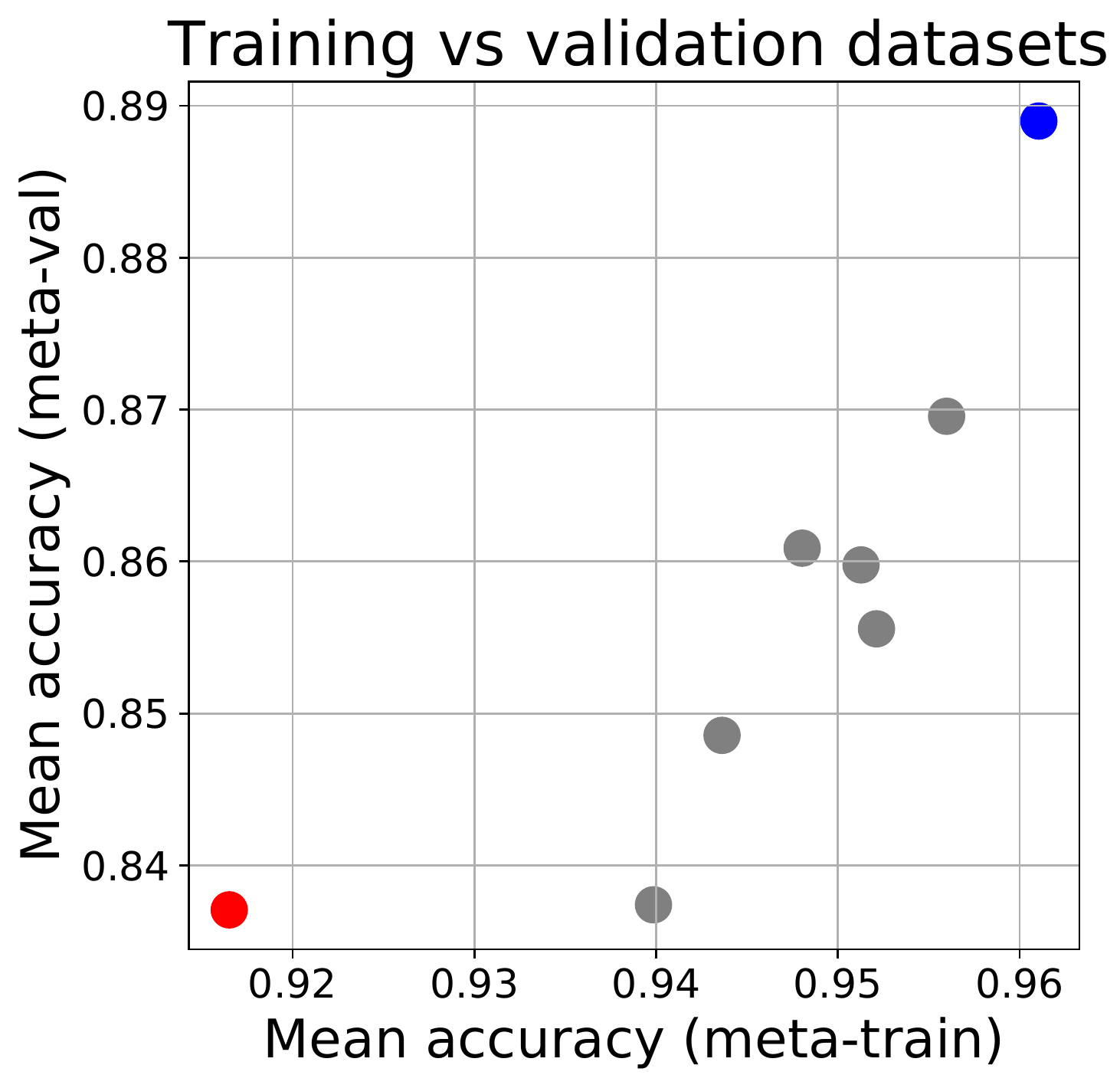} & %
    \includegraphics[width=0.41\textwidth]{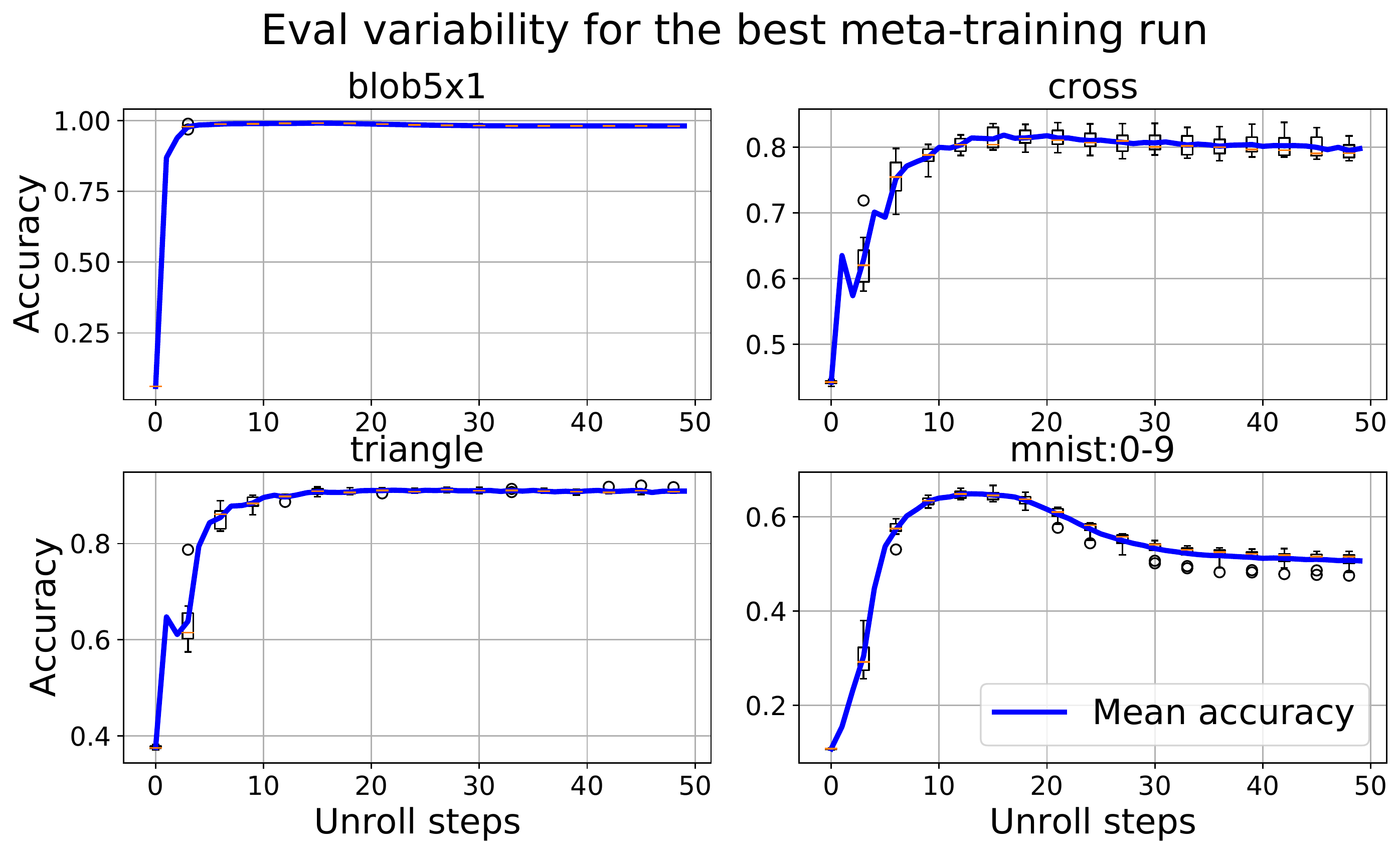} %
    \end{tabular}
    \caption{Meta-learning on toy datasets.  \emph{Top row:} the validation performance on training datasets. The spread indicates variation across multiple meta-training runs. \emph{Bottom left:} shows the variation on validation datasets. \emph{Bottom right:} the variation within a single run, but fora different synapse initializations. \emph{Bottom middle:} the correlation between mean meta-validation and meta-training accuracy across different runs at unroll 10. \emph{For all the graphs} the blue line shows the run with the highest average meta-training accuracy and the red one the lowest. Since genomes are selected based on their \emph{average} meta-training, it is expected that for some tasks they are not the worst.}
    \label{fig:metalearning-on-toys}
\end{figure*}
In this section we describe experimental evaluation of update rules using \sourdough. Our code uses tensorflow~\cite{tensorflow2015-whitepaper} and JaX~\cite{jax2018github} libraries. All our experiments run on GPU. Typically a single genome can be trained on a single GPU in between 30 minutes and 20 hours depending on configuration. For some experiments we run multiple identical runs to estimate variance.
For all our experiments we use the same set of basic parameters as described in Appendix \ref{sec:exp_params}. In section \ref{sec:ablation:study} we study several alternative formulations to show their impact on convergence and stability. Code for the paper is available at \url{https://github.com/google-research/google-research/tree/master/blur}

\subsection{Meta-learning simple functions}

We first consider a very simple setup where we try to learn toy-examples with two variable inputs. The datasets are shown in Fig.~\ref{boolean-datasets}.  The tasks we use for training are \texttt{and}, \texttt{xor}, \texttt{two-moon}~\cite{scikit-learn} and several others. We then validate  it on held out datasets shown in Fig.~\ref{boolean-datasets}. These datasets include both in-domain (e.g.\ other 2-class functions), five-class function of two variables {\it blobs}~\cite{scikit-learn}
and out-of-domain MNIST~\cite{mnist}. We train a two-layer network using a two-state genome, and use the same two-layer architecture for meta-training and meta-validation. The results are shown in Fig.~\ref{fig:metalearning-on-toys}.

\begin{figure*}[ht]
    \centering
    \includegraphics[width=0.85\textwidth]{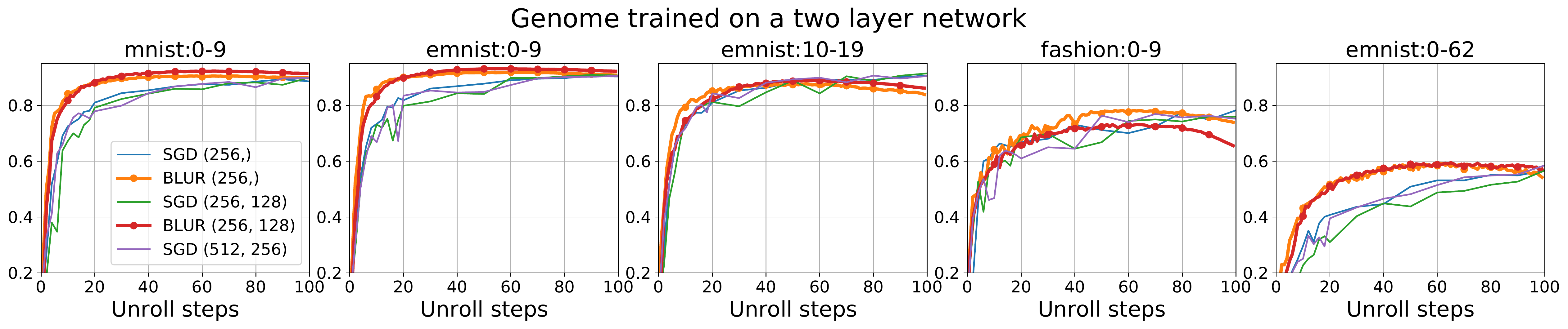}
    \includegraphics[width=0.13\textwidth]{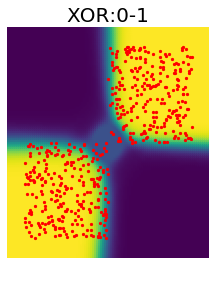}
    \caption{Generalization of a genome meta-trained with 2-layer/4-state architecture on MNIST to other datasets. Note that our search only explored up to 50 unroll steps. Our method converges much faster than SGD in the explored training trajectory, however after about 100 steps SGD reaches the same accuracy and continues to grow. The rightmost visualization applies the same genome to learn a \texttt{xor} function. }
    \label{fig:mnist-generalization}
\end{figure*}
\label{meta-learning-mnist}
\begin{figure}[ht]
\includegraphics[width=0.45\textwidth]{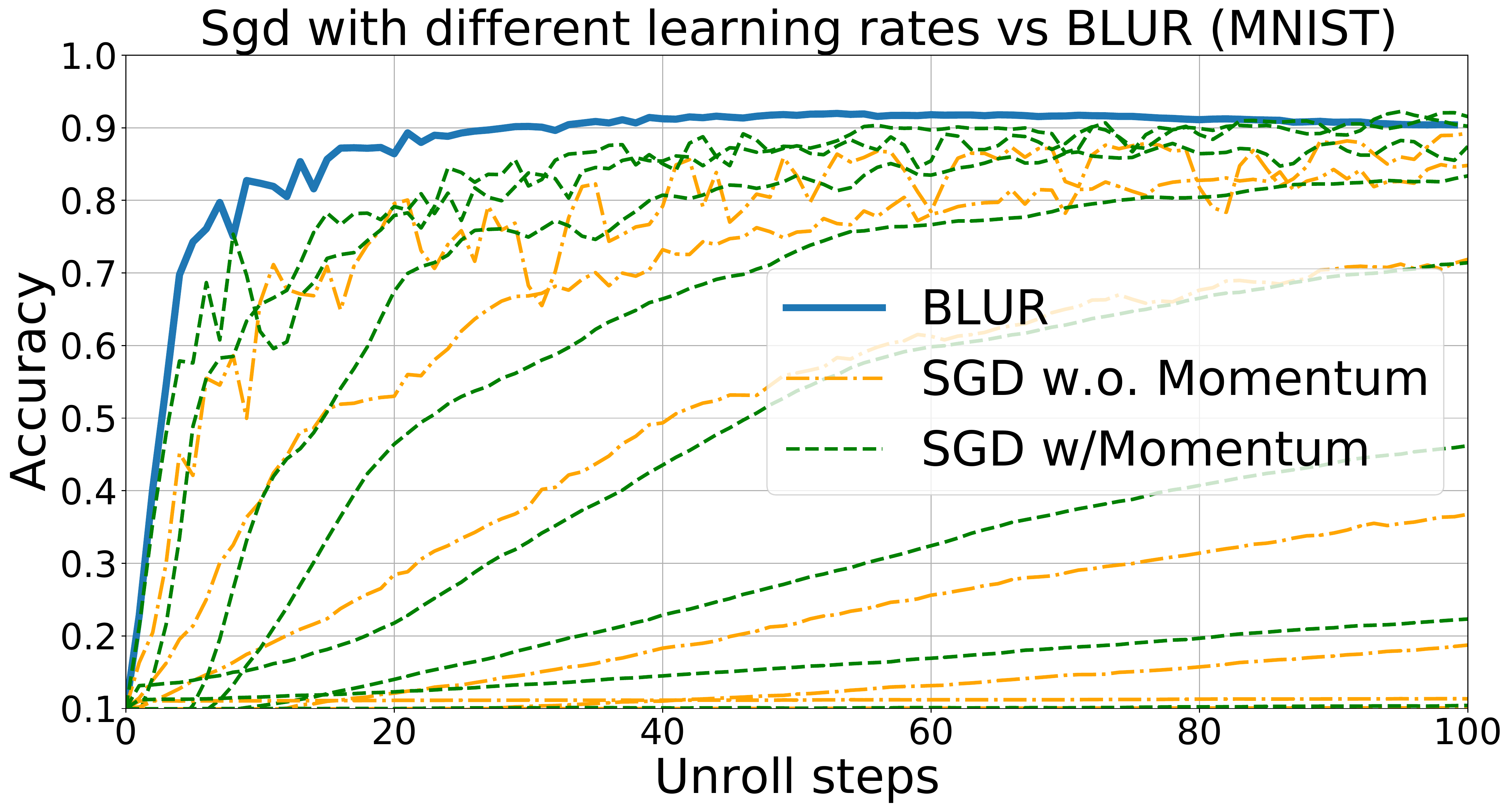}
\caption{Performance of 4-state BLUR network vs. SGD with different learning rates.}
\label{fig:blur_vs_sgd}
\end{figure}
\begin{figure}[ht]
    \centering
    \includegraphics[width=0.23\textwidth]{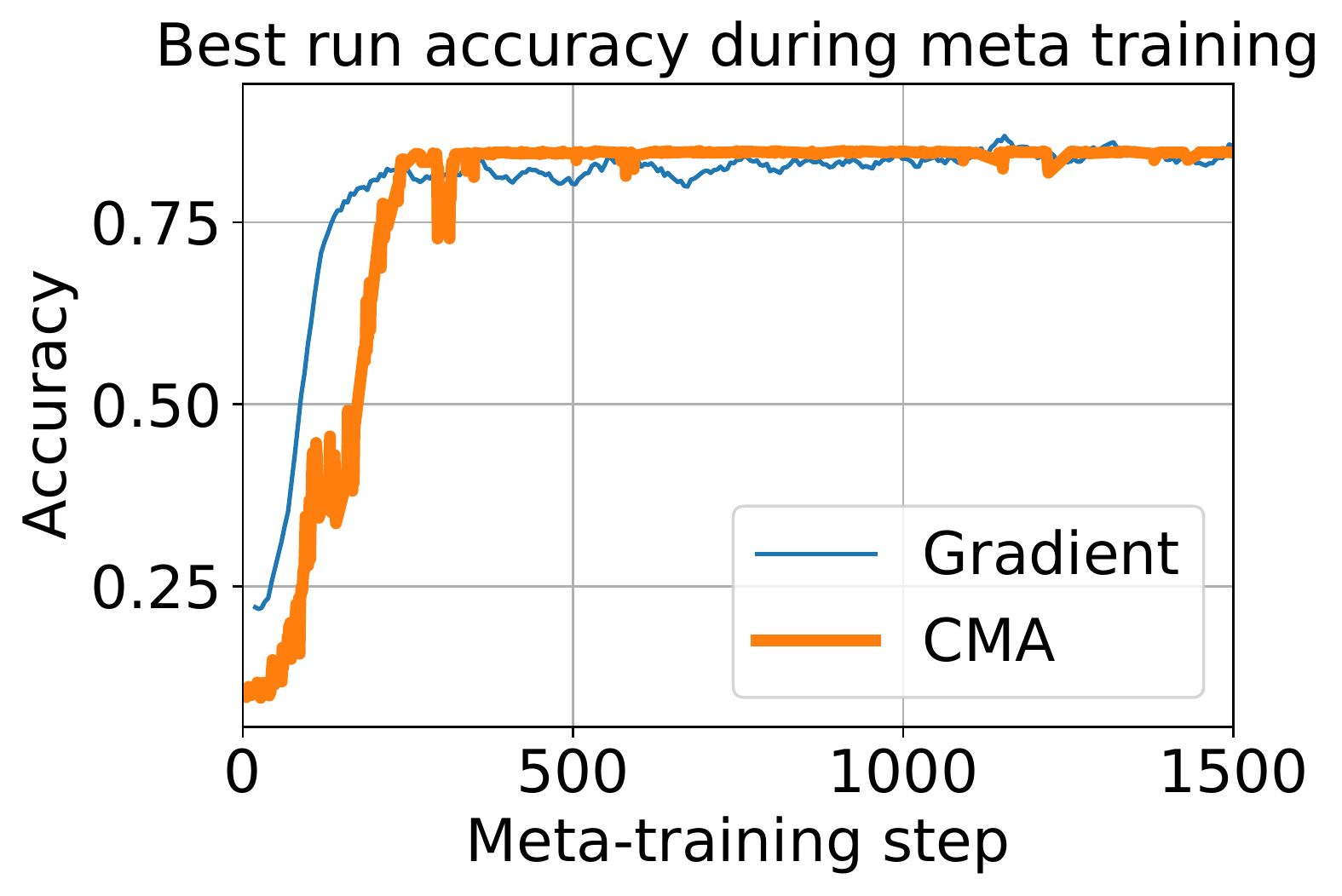}
    \includegraphics[width=0.23\textwidth]{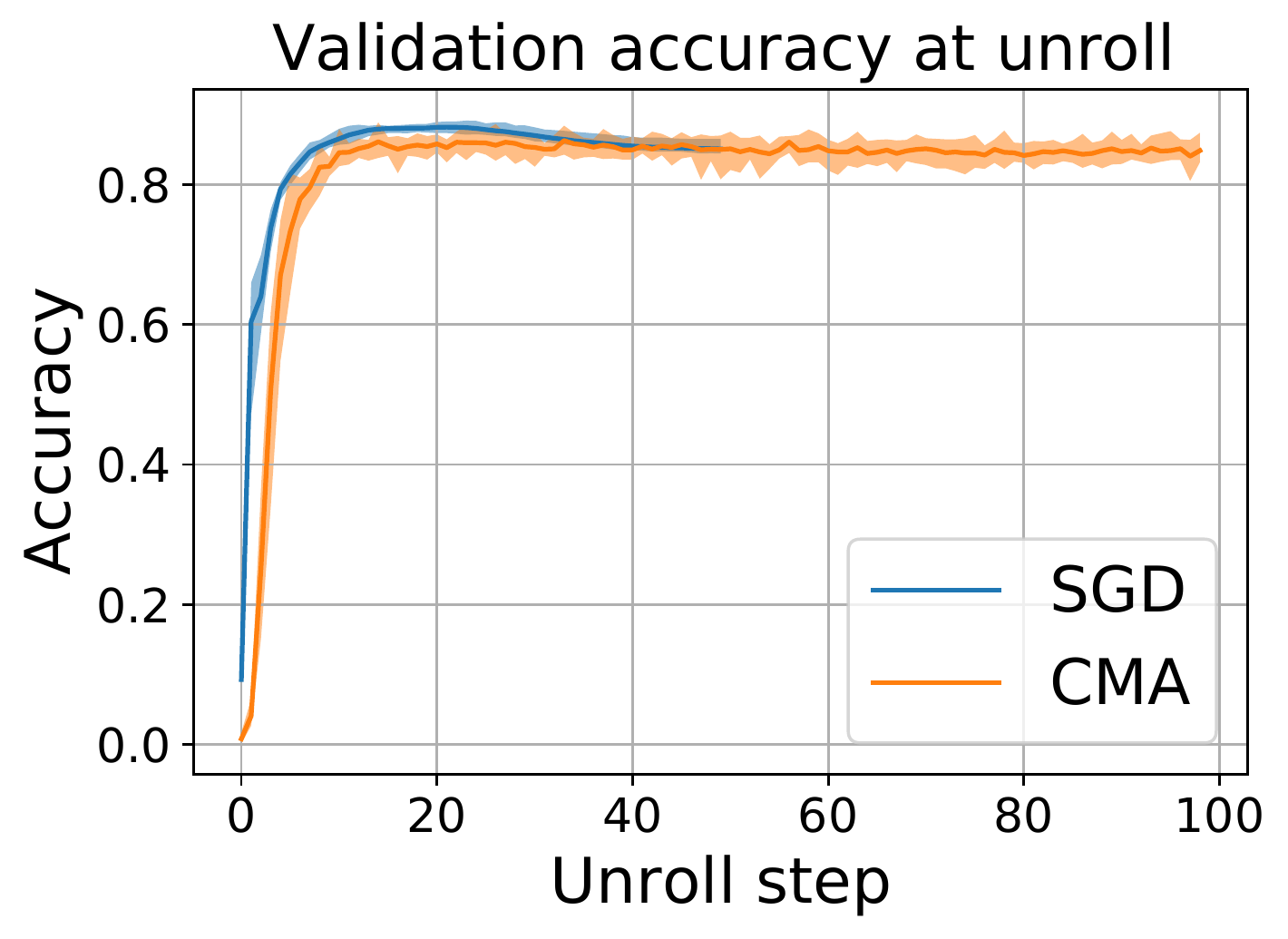}
    \caption{Using CMA to learn best non-differentiable accuracy objective, against gradient based meta training. \emph{Left:} meta-training trajectory \emph{right:} fully trained genome learning MNIST. Both methods show best run out of 8.}
    \label{fig:cma_metalearning}
\end{figure}

\begin{figure}[t]
    \centering
    \includegraphics[width=0.48\textwidth]{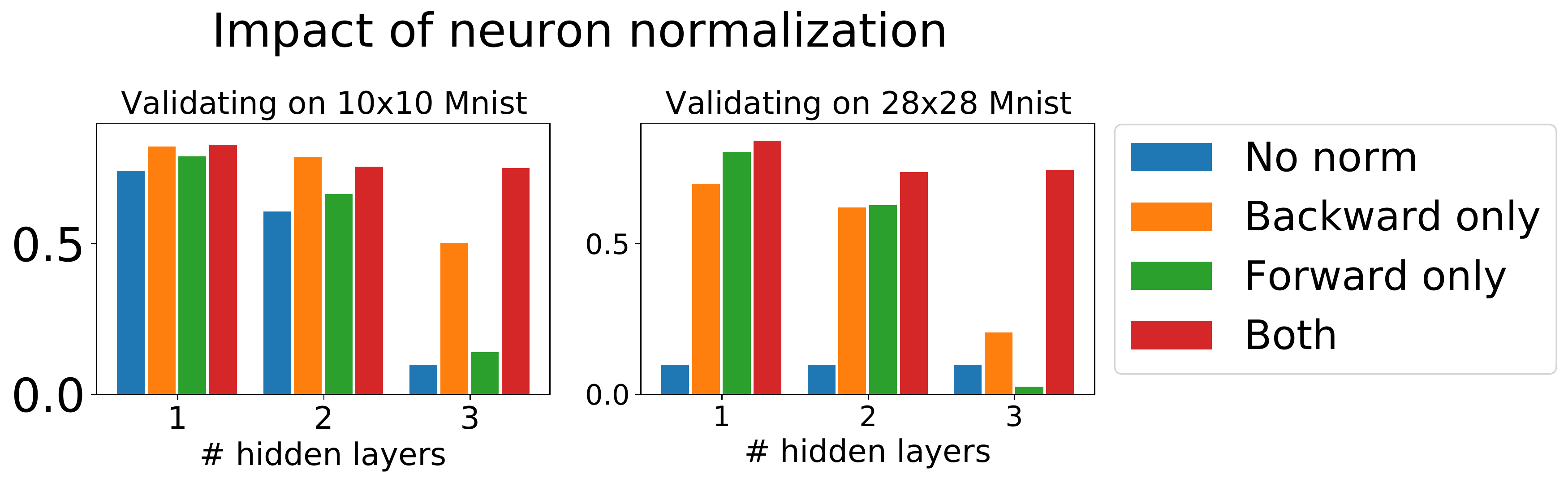}
    \caption{Importance of neuron normalization.  The genomes that do not use normalization produced did not generalize as well to different resolutions and deeper networks. }
    \label{fig:normalization}
\end{figure}
\subsection{Generalization capabilities}
In this section we explore the ability of genomes that we found to generalize to new datasets. We use MNIST as a meta-training dataset. Since MNIST generally requires more than 10 steps to converge we use a variant of curriculum training to gradually increase the number of unrolls. We start with 8-identical randomly initialized genomes and train them for 10,000 steps with 10 unrolls. Then we increase the unroll number by 5 for each consecutive 10,000 steps and synchronize genomes across all runs. The example of meta-training training accuracy is shown \supplementalref{fig:mnist-metatraining}. In Fig.~\ref{fig:mnist-generalization} we show the performance of a curriculum-trained genome. The genome was trained on 3 tasks: full MNIST, and two half-size MNIST datasets, one with digits from 0 to 4 and another from 5 to 9. Interestingly, even this naive setup that uses just three slightly different tasks shows improved generalization abilities. For meta-training we used the MNIST dataset cropped to 20x20 and resized to 10x10.  For meta-validation we always used 28x28 datasets. Specifically we used MNIST, a 10-class letter subset of
E-MNIST~\cite{emnist}, Fashion MNIST~\cite{fashion}, and the full 62-category E-MNIST. This shows that the meta-learned update rules can successfully learn unrelated tasks without ever having access to gradient functions. We didn't include the graph of evaluation accuracy for \emph{meta-training} variant of  MNIST that used cropped and down-sampled digits, but we note  that it generally produced results that were about 1\% below 28x28 MNIST, thus suggesting excellent meta-generalization.

Finally, on the same Figure~\ref{fig:mnist-generalization} we show an example of an MNIST-trained genome applied to an out-of-domain Boolean task. 

\subsection{Comparison with SGD}
In figure \ref{fig:blur_vs_sgd} we compare the convergence performance on MNIST dataset of \sourdough vs. SGD with and without momentum with different values of learning rate spanning 4 orders of magnitude.

\subsection{Training using evolution strategies}
\label{sec:cma} 
Using the evolution strategies to train genomes are appealing for two reasons. First, it enables us to train networks with a much larger number of unrolls. Second, it allows us to find genomes with non-differentiable objectives.

In this section we use a simplified setup without curriculum learning.  We train a network with a single hidden layer of 512 channels and 2 states on MNIST downsampled to 14x14. Each meta-learning step represents training the network on 15 batches of 128 inputs, then evaluating its accuracy on 20 batches of 128 inputs. We use the CMA-ES/pycma~\cite{hansen2019pycma} library to optimize the network genome with respect to the (non-differentiable) 20 batch accuracy. Using a population of 80 parallel experiments per time step, CMA-ES is able to learn a genome that achieves 0.84 accuracy in 400 steps. Accuracy as a function of meta-learning steps can be seen in Fig.~\ref{fig:cma_metalearning}. Evaluating the generalization of this genome over different numbers of unrolls reveals that it reaches full accuracy in 15 unroll steps as expected, and plateaus without significant decay and results in comparable performance as gradient-based search.

\subsection{Ablation study}
\label{sec:ablation:study}
In this section we explore the importance of several critical parameter choices. We will be training with the same setup as in section \ref{meta-learning-mnist}, but instead of verifying it on the meta-validation dataset, we will measure the impact on MNIST.

\paragraph{Normalization} Normalization plays a crucial role in the stability of our meta-training process and improves the final training accuracy. Fig.~\ref{fig:normalization} shows a meta-training accuracy comparison of identical runs with forward/backward normalization turned off.

\paragraph{Impact of non-linearity} In contrast with chain-rule back-propagation, our learning algorithm uses symmetric non-linearity for simplicity. Curiously, it appears that the independent choice of non-linearity on the forward and backward pass has relatively little impact on our ability to find genomes that can learn.  Generally the space seems to be insensitive to the choice of non-linearity used on either forward or backward pass. In Appendix we include a table showing the variation in validation accuracy across different non-linearities.

\begin{figure}
    \centering
    \begin{tabular}[c]{@{\hspace{0.0\linewidth}}c|@{\hspace{0.01\linewidth}}c@{\hspace{0.01\linewidth}}c@{\hspace{0.0\linewidth}}c@{\hspace{0.0\linewidth}}}
    & & \multicolumn{2}{c}{Meta-train} \\\hline
    & & \hspace{2ex} MNIST & \hspace{2ex} 10-way Omniglot \\
    \multirow{2}{*}{\rotatebox{90}{\hspace{9ex}Meta-Eval}} 
    & \rotatebox{90}{\hspace{6ex}MNIST} & \includegraphics[width=0.21\textwidth]{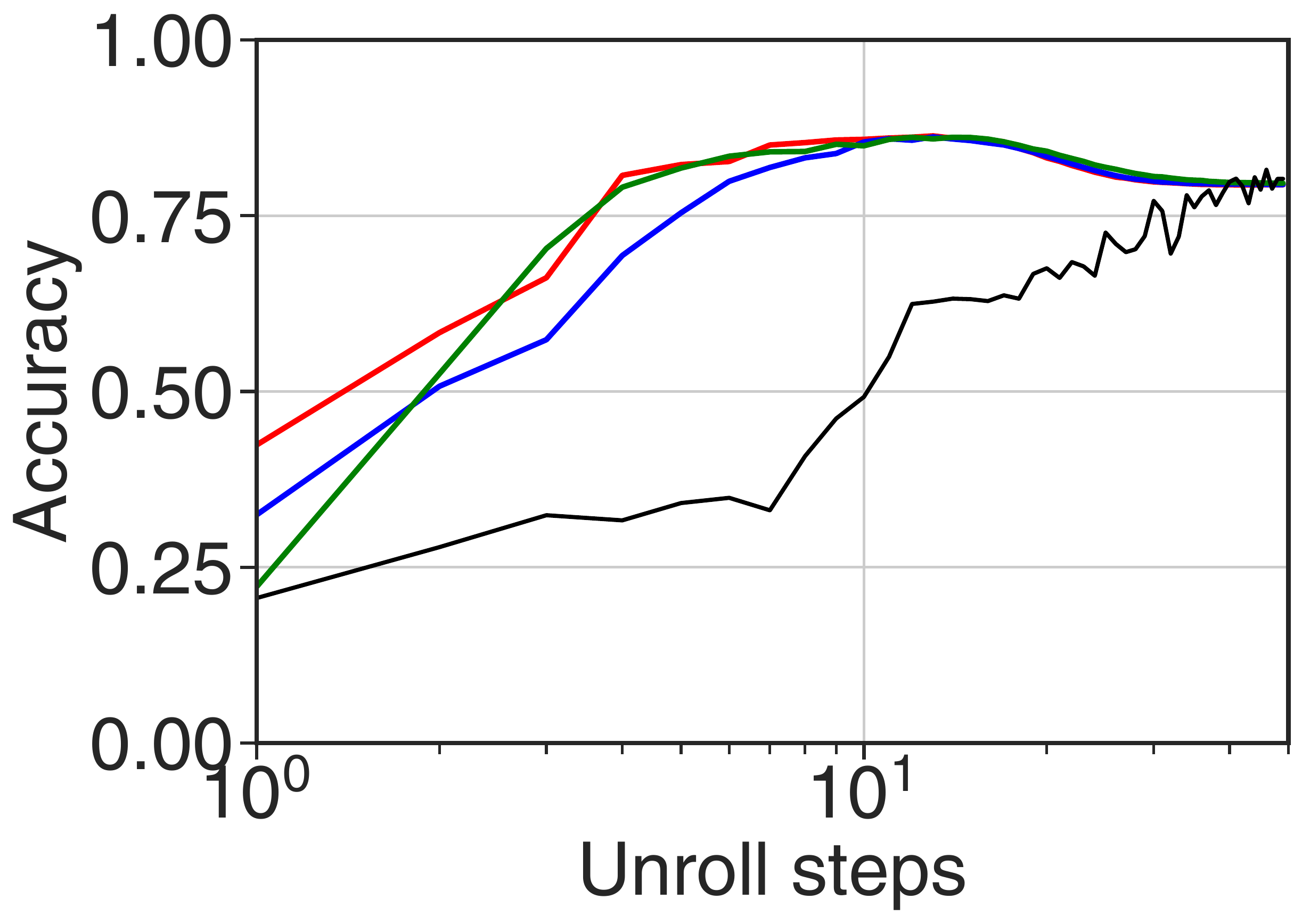} &
    \includegraphics[width=0.21\textwidth]{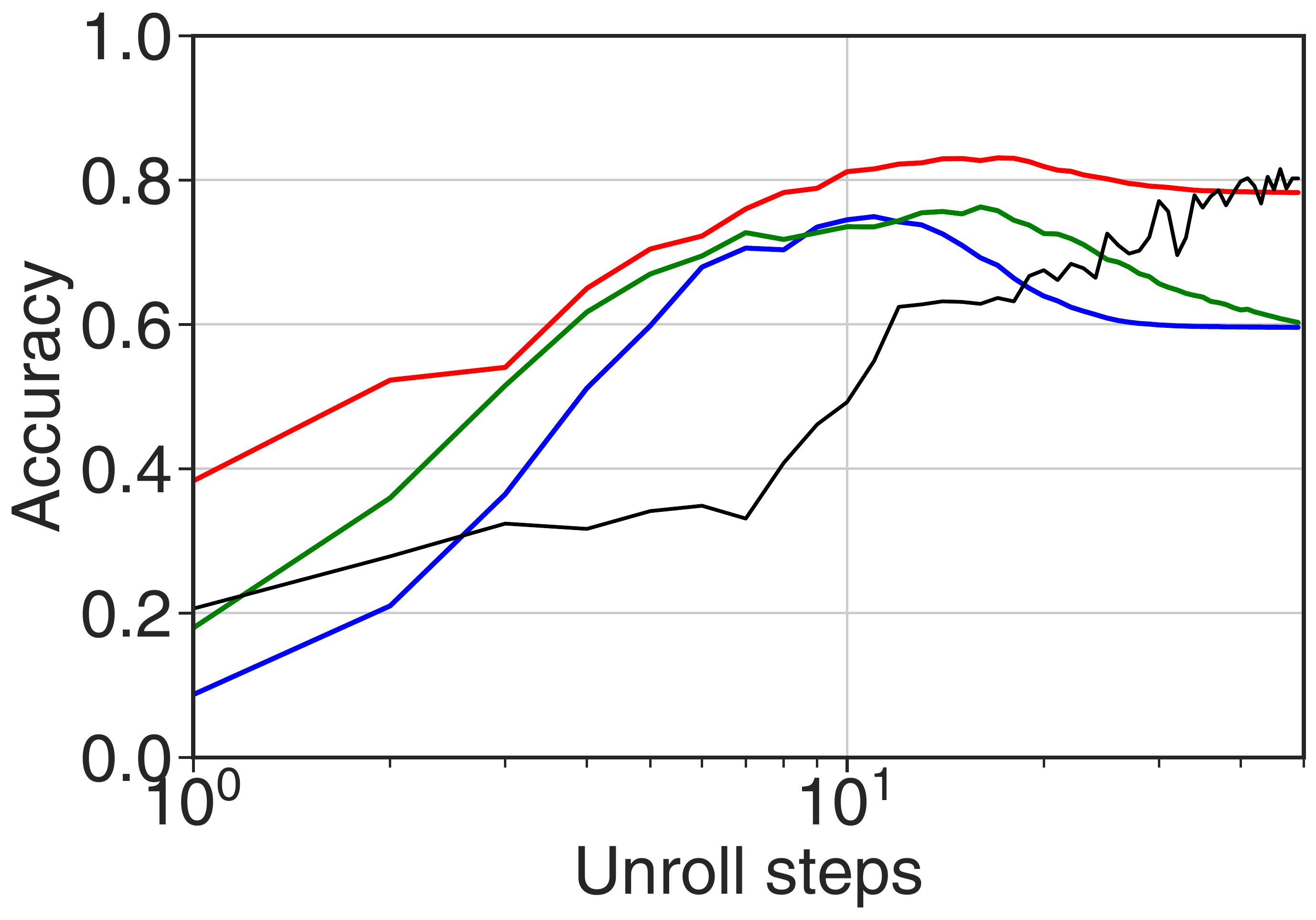} \\
    & \rotatebox{90}{\hspace{6ex}Omniglot} & \includegraphics[width=0.21\textwidth]{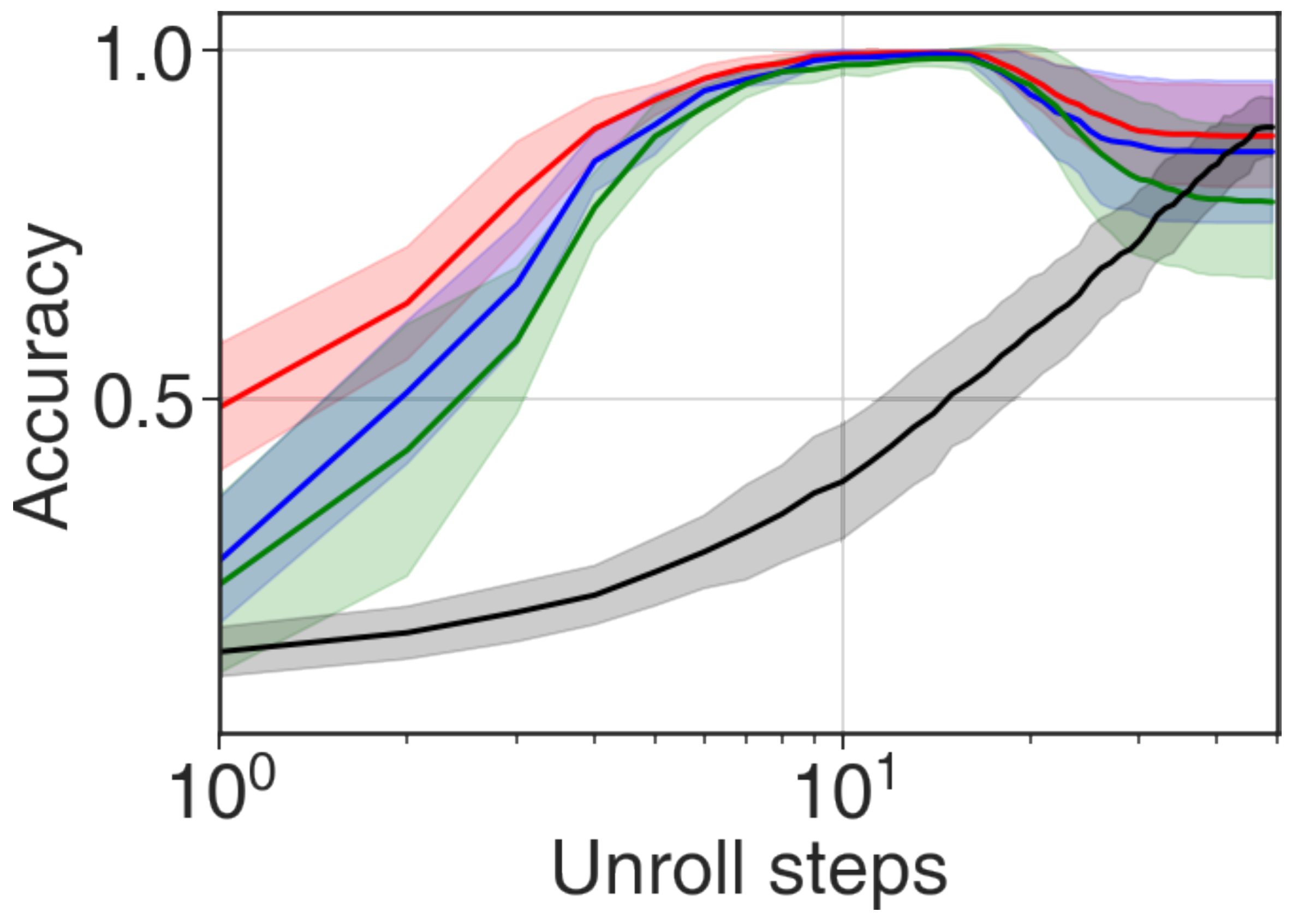} & 
    \includegraphics[width=0.21\textwidth]{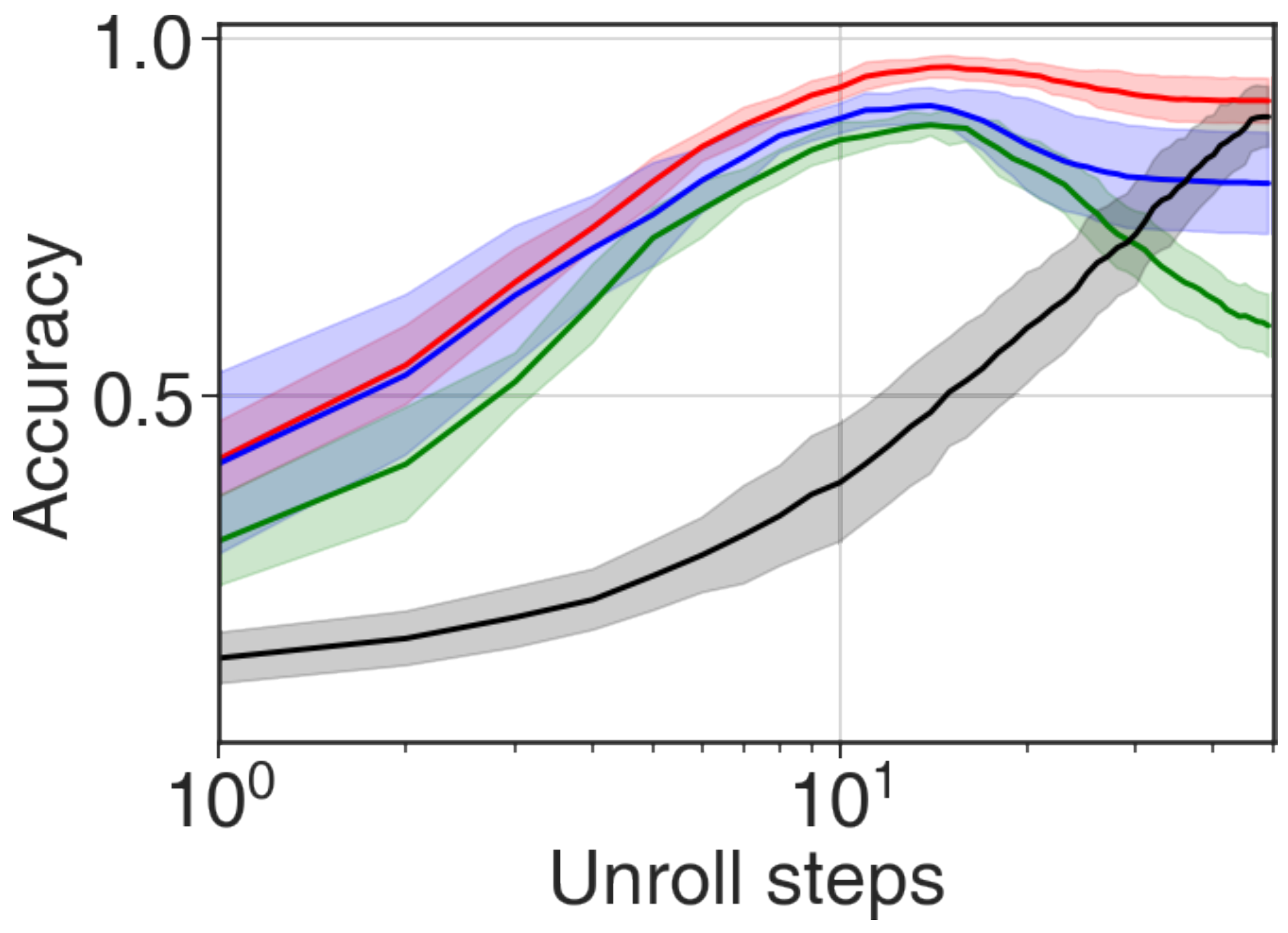}
    \end{tabular}
    \includegraphics[width=0.45\textwidth]{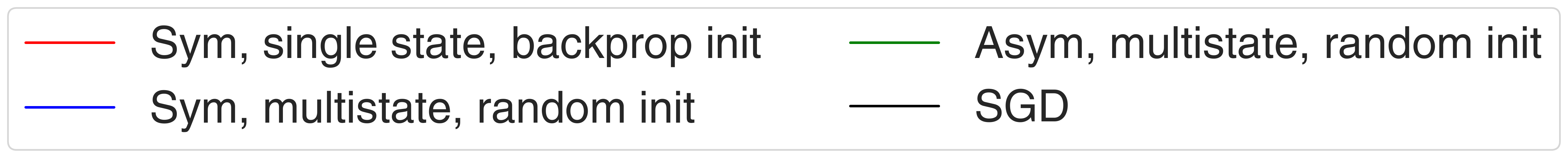}
    \caption{Accuracy of different variants of \sourdough trained on MNIST (left) or 10 episodes from Omniglot (right) to 10 unrolls and evaluated on MNIST (top) and Omniglot (bottom). Black line corresponds to the SGD. Errorbars show standard deviation of the accuracy over 10 different subsets of Omniglot. Only the best result of 8 runs is plotted.}
    \label{fig:compare_flags_omniglot}
\end{figure}

\paragraph{Symmetry of the synapses} Here we compare our framework across combinations of three different dimensions: (a) using symmetric or asymmetric synapses for backward and forward passes, (b) using single or multi-state synapses and (c) initializing genome close to backpropagation (i.e.\ using $\nu = \left(\begin{smallmatrix} 1 & 0 \\ 1 & 0 \end{smallmatrix} \right)$, $\mu = \left(\begin{smallmatrix} 1 & 0 \\ 0 & 1 \end{smallmatrix} \right)$, $\tilde\nu = \left(\begin{smallmatrix} 1 & 0 \\ 0 & 1 \end{smallmatrix} \right)$, $\tilde\mu = \left(\begin{smallmatrix} 0 & 1 \\ 1 & 0 \end{smallmatrix} \right)$ as defined in Sec.~\ref{sec:backprop_genome}). We trained eight different combinations above on the MNIST dataset and on a 10-way episode learning from Omniglot (at every meta-iteration sampling 10 different random classes from the 1200 class subset of Omniglot). We then evaluate the resulting genomes on test set of MNIST or 10 different subsets of Omniglot tasks that were not part of the training set. In Fig.~\ref{fig:compare_flags_omniglot} we show three best variants of these parameters (the other five had much worse results and available in the Appendix) as well as comparison with SGD trained using cross entropy loss and a learning rate $0.001$. We noticed that while having asymmetric forward and backward synapses allows our system to be strictly more general, it does not always lead to better generalization. The simplest variant with symmetric, single-state synapses tends to be the winner, however it has to be initialized from the backprop genome. Other two variants that performed well: symmetric, multi-state genomes and the most general asymmetric, multi-state genome.

\begin{figure}
    \centering
    \includegraphics[width=0.49\textwidth]{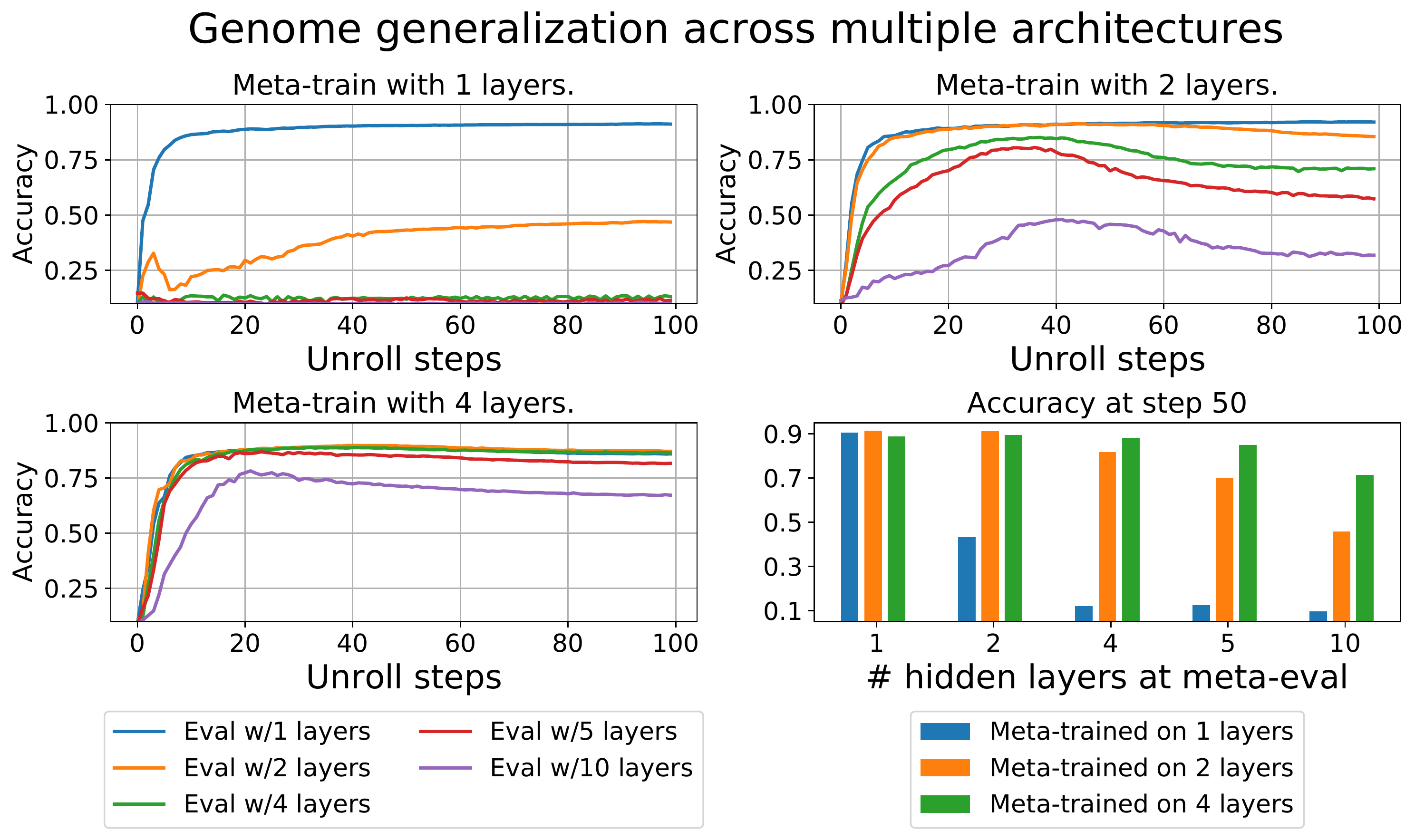}
    \caption{Genome generalization across architectures. Here we show that the genomes that were trained using deeper architectures work well on shallower architectures but not vice versa.}
    \label{fig:deeper-wider-networks}
\end{figure}

\paragraph{Learning genomes for deeper and wider networks}
In this experiment we meta-train our genome on networks that contain 2, 3, and 4 hidden layers and explore their ability to generalize across changes in layer sizes and architecture, as measured by their performance on MNIST after 10 steps. We use the same setup as in section \ref{meta-learning-mnist}.
The results are shown in Fig.~\ref{fig:deeper-wider-networks}. Interestingly we discover that  genomes generalize from more complex architectures to less complex architectures, but not vice versa! For instance genomes trained using a 2-layer network performed well when tasked to use a single-hidden layer. However they diverged when training a 4-layer network. A 4-layer genome was able to train networks both with shallower (e.g.\ 1 to 3 layers) and deeper (10) architectures.

\section{Conclusions and Future Work}
    In this work, we define a general protocol for updating nodes in a neural network, yielding a domain of ``genomes'' describing many possible update rules, of which gradient descent is one example. Useful genomes are identified by training networks on training tasks, and then their generalization is evaluated on unseen tasks. We have shown that it is possible to learn an entirely new type of neural network that can be trained to solve complex tasks faster than traditional neural networks of equivalent size, but without any notion of gradients. Our approach can be combined with many existing model representations with differentiable or non-differentiable components. 
    
    There are many interesting directions for future exploration. Perhaps, the most important one is the question of scale. Here one intriguing direction is the connection between the number of states and the learning capabilities. Another possible approach is extending the space of update rules, such as allowing injection of randomness for robustness, or providing an ability for neurons to self-regulate based on current state. Finally the ability to \emph{extend} existing genomes to produce ever better learners, might help us scale even further. Another intriguing direction is incorporating the weight updates on both forward and backward passes. The former can be seen as a generalization of unsupervised learning, thus merging both supervised and unsupervised learning in one gradient-free framework.
\bibliography{main}
\bibliographystyle{icml2021}
\clearpage

\appendix
\begin{figure*}[t]
    \centering
    \begin{tabular}[c]{@{\hspace{0.0\linewidth}}c|@{\hspace{0.01\linewidth}}c@{\hspace{0.0\linewidth}}}
    EMNIST Subtasks (28x28) & MNIST (28x28) \\
    \includegraphics[width=0.7\textwidth]{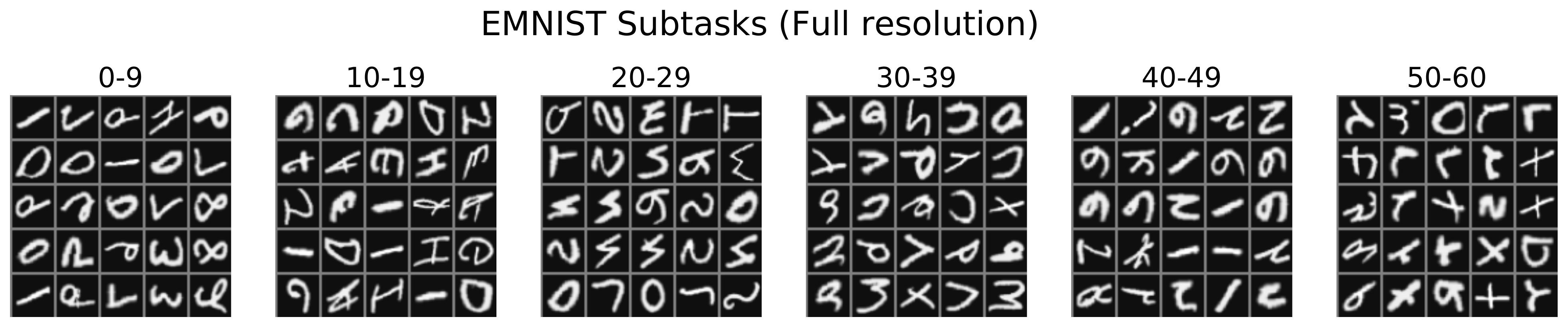} & 
    \includegraphics[width=0.115\textwidth]{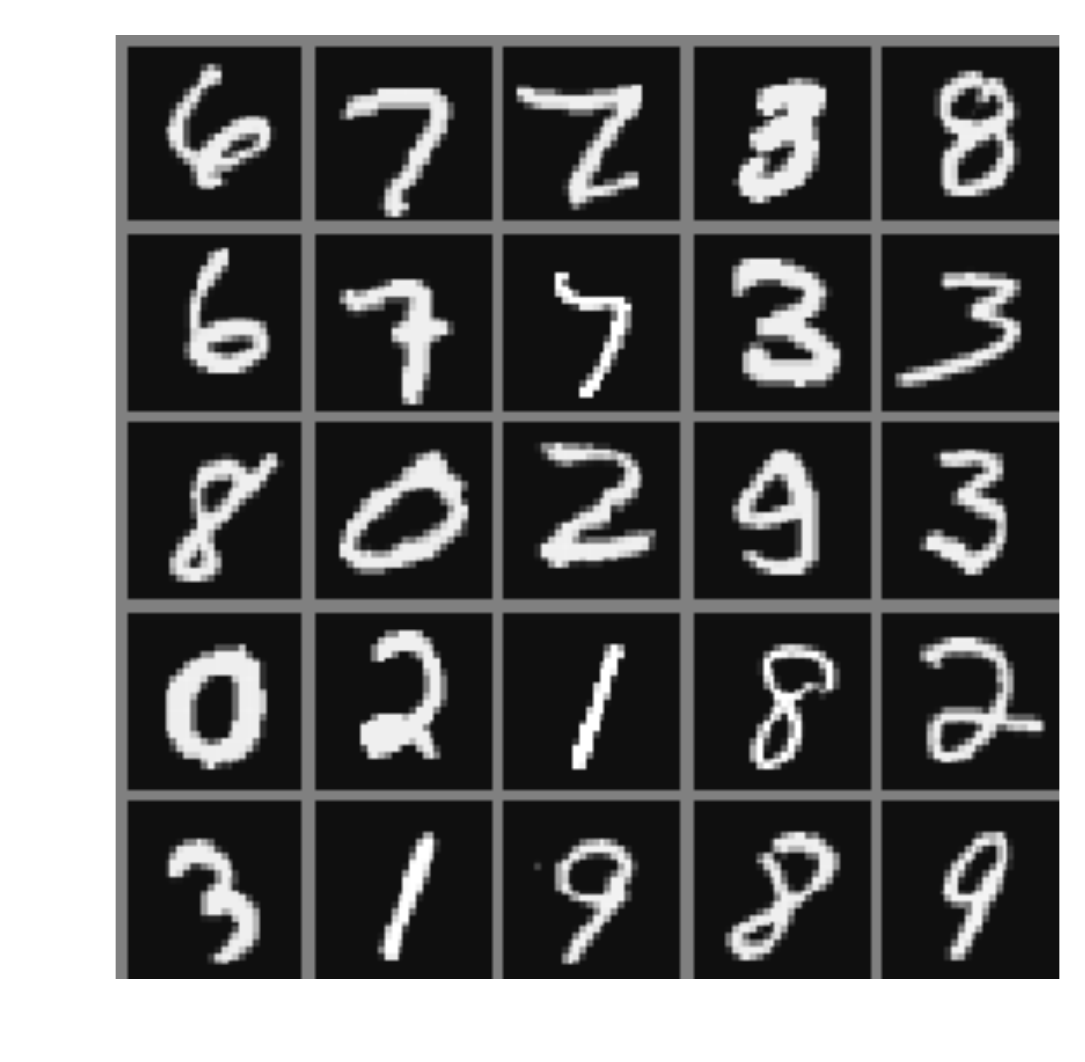}
    \\ 
    EMNIST Subtasks (10x10) & Fashion MNIST (28x28) \\
    \includegraphics[width=0.7\textwidth]{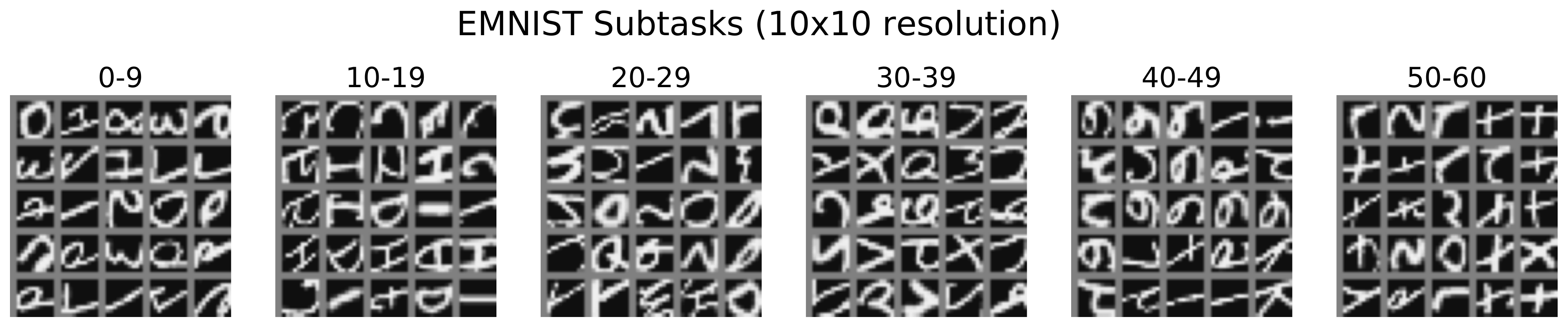} & 
    \includegraphics[width=0.115\textwidth]{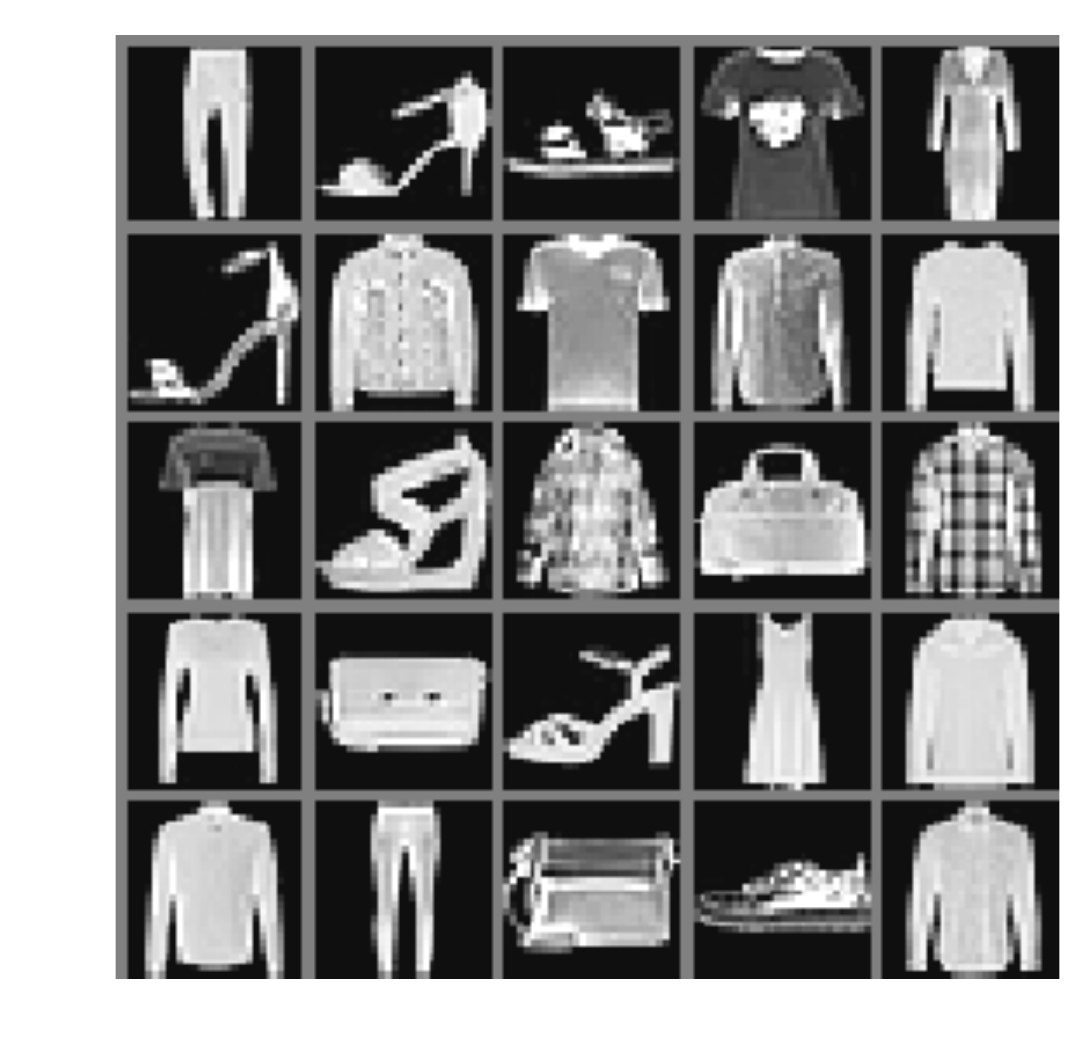}
    \end{tabular}
    \caption{Dataset subtasks at 10x10 and 28x28 resolution}
    \label{fig:emnist-subtests}
\end{figure*}

\section{Connection to Gradient Descent}
    \label{sec:as_sgd_ext}
    Instead of verifying the equivalence of our update rule $\Delta \vec{w}(\vec{w})$ to GD (with $\vec{w}$ including both forward and backward weights), we may ask a more general question of equivalence to a generalized gradient descent satisfying:
\begin{equation}
    \label{eq:metric_dw}
    \Delta w_j = -\gamma \sum_i g^{-1}_{ij} \pd{L_{\rm equiv}}{w_i},
\end{equation}
where $g_{ij}$ are components of a Riemannian metric tensor $\tensor{g}(\vec{w})$ and $\vec{w}$ are the model weights.
Since Eq.~\eqref{eq:metric_dw} can be rewritten as $-\gamma dL_{\rm equiv} = \tensor{g} \Delta \vec{w}$, where $dL_{\rm equiv}$ is the exterior derivative of $L_{\rm equiv}$, it follows that $d(\tensor{g} \Delta \vec{w}) = 0$, or equivalently
\begin{equation}
    \label{eq:metric_nec}
    \pd{}{w_r} \left( \sum_j g_{ij} \Delta w_j \right) = \pd{}{w_i} \left( \sum_j g_{rj} \Delta w_j \right).
\end{equation}
For contractible parameter spaces, this condition is also sufficient for the existence of $L_{\rm equiv}$ \cite{lee2013smooth}.

\def\tS{\tensor{\Sigma}}
\def\tA{\tensor{A}}
\def\tR{\tensor{R}}
\def\tQ{\tensor{Q}}
\def\tg{\tensor{g}}
\def\tG{\tensor{\Gamma}}
\def\tZ{\tensor{Z}}
\newcommand{\pld}[2]{\partial #1 / \partial #2}

\paragraph{Constant $\tg$.}
Consider a special case where $\tg$ is constant, i.e., independent of $\vec{w}$.
Then we can rewrite Eq.~\eqref{eq:metric_nec} as
\begin{gather*}
  \sum_j g_{ij} \pd{\Delta w_j}{w_r} = \sum_j g_{rj} \pd{\Delta w_j}{w_i},
\end{gather*}
or introducing an $n\times n$ matrix $Q_{ij} := \pld{\Delta w_i}{w_j}$ with $n=\dim \vec{w}$ as:
\begin{gather}
	\label{eq:cond}
	\tZ[\tg,\tQ] := \tg \tQ - (\tg \tQ)^\top = 0.
\end{gather}
For $\tg$ to be a metric tensor, it has to be symmetric and positive definite and Eq.~\eqref{eq:cond} should be satisfied for all possible $\tQ(\vec{w})$ calculated at all accessible states $\vec{w}$.

In our numerical experiments, we studied a pre-trained two-state model learning a 2-input Boolean task with a single hidden layer of size $5$.
Finding a null space of the matrix corresponding to a linear system~\eqref{eq:cond} for some $\tQ(\vec{w}_*)$, we then checked the validity of Eq.~\eqref{eq:cond} for the basis vectors of this null space and $100$ other $\tQ(\vec{w})$ matrices calculated at different other random states $\vec{w}$.
Out of $265$ basis vectors\footnote{Found by performing SVD and taking vectors corresponding to singular values below a $10^{-8}$ threshold.}, $4$ solved Eq.~\eqref{eq:cond} for all $100$ of matrices $\tQ$, but as we verified later, none of those $4$ vectors (or $\tg$ candidates) were positive-definite and actually had $0$ eigenvalues.
This shows that at least in this particular case, our update rules $\Delta \vec{w}(\vec{w})$ cannot be represented as a generalized GD as shown in Eq.~\eqref{eq:metric_dw}.

\paragraph{Equivalence to SGD and closed trajectories.}
Solving Eq.~\eqref{eq:metric_nec} for a Riemannian metric $\tensor{g}$ in a more general case is difficult.
There are certain special cases, however, where Eq.~\eqref{eq:metric_nec} does not have solutions.
For example, if the vector field defined by $\Delta \vec{w}$ supports closed integral curves, there are trivially no $\tensor{g}$ and $L_{\rm equiv}$ that would satisfy Eq.~\eqref{eq:metric_dw}.
Indeed, if they existed, then parameterizing this closed integral curve $\Gamma$ by $t\in [0,1]$, we would obtain:
\begin{multline*}
    0 = \int\limits_0^1 \frac{dL_{\rm equiv}(\Gamma(t))}{dt} \, dt =
    \int_0^1 dL_{\rm equiv}(\dot{\Gamma}) \, dt = \\ =
    - \gamma \int_0^1 \tensor{g}^{-1} dL_{\rm equiv} dL_{\rm equiv} \, dt = 0,
\end{multline*}
which cannot hold for a Riemannian metric $\tensor{g}$.

\section{Modified Oja's Rule}
    \label{sec:oja_ext}
    
Oja's rule is generally derived by augmenting weight update with the normalization multiplier \cite{oja1982simplified}.
Writing a normalized weight update as:
\begin{gather}
    \label{eq:orig_upd}
    w^*_{ij} =
    \frac
      {w_{ij} + \gamma \phi[w_{ij}, x_i, y_j]}
      {\left[ \sum_i \left( w_{ij} + \gamma \phi[w_{ij}, x_i, y_j] \right)^2 \right]^{1/2}},
\end{gather}
where $x$ and $y$ are pre-synaptic and post-synaptic activations, and $w^*$ is the weight at the next iteration, we can see that in the limit of $\gamma \to 0$, this update rule can be rewritten as:
\begin{multline}
    \label{eq:oja_phi}
    w^*_{ij} =
      w_{ij} + \gamma \phi[w_{ij}, x_i, y_j] - \\ - \gamma w_{ij} \sum_i w_{ij} \phi[w_{ij}, x_i, y_j] + O(\gamma^2).
\end{multline}

For the Hebbian update $\phi = x_i y_j$, the last $O(\gamma)$ term in Eq.~\eqref{eq:oja_phi} becomes a familiar Oja's term:
\begin{gather}
    \label{eq:oja_canon}
    - \gamma w_{ij} \left(\sum_r w_{rj} x_r\right) y_j =
    - \gamma w_{ij} y_j^2.
\end{gather}

For more complicated networks and families of update rules, the canonical form of the Oja's rule \eqref{eq:oja_canon} is no longer applicable.
Let us derive a saturating term for the family of update rules from Section 3.2, where the updated weight now has the form $\tensor{w}^* = \tilde{f} \tensor{w} + \tilde{\eta} \tensor{\phi}$.
Expressing $\tilde{f}$ as $1 + \tilde{\eta} \beta$, we can rewrite the update rule as $\tensor{w}^* = \tensor{w} + \tilde{\eta} \tensor{\phi}_\circ$, where $\tensor{\phi}_\circ = \tensor{\phi} + \beta \tensor{w}$ and the corresponding saturating term in the update rule then becomes
\begin{gather*}
    - (\tilde{f} - 1) w_{ij}^c \sum_{r} (w_{rj}^c)^2
    - \tilde{\eta} w_{ij}^c \sum_{r} w_{rj}^c \phi[\vec{w}_{rj}, \vec{a}_r, \vec{a}_j],
\end{gather*}
where both activations $\vec{a}$ and weights $\tensor{w}$ now also have an additional ``state'' dimension.
Substituting $\phi = \sum_{e,d} a_{i}^e \tilde{\nu}^{ec} \tilde{\mu}^{cd} a_{j}^d$ here, we finally obtain the following inhibitory component of the update rule:
\begin{multline*}
    (\Delta w_{ij}^c)^{\rm Oja} = - (\tilde{f} - 1) w_{ij}^c \sum_{r} (w_{rj}^c)^2 - \\ -
    \tilde{\eta} w_{ij}^c \sum_{r,e,d} w_{rj}^c a_r^e \tilde{\nu}^{ec} \tilde{\mu}^{cd} a_{j}^d.
\end{multline*}

\section{Additional Experiments}
    \subsection{Parameters}
\label{sec:exp_params}
\begin{tabular}{|l|c|l|c|}
\hline
\multicolumn{2}{|c|}{Meta optimizer}& \multicolumn{2}{c|}{Genome}\\
\hline
Optimizer  & Adam  & States per neuron & 2-8\\
LR        & 0.0005 & Batch size & 128 \\ 
Gradient clip & 10 & Unroll length & 10-50 \\    
\hline
\end{tabular}

Fig.~\ref{fig:emnist-subtests} shows a sample from datasets used in the experiments.

\begin{figure}
    \centering
    \includegraphics[width=0.45\textwidth]{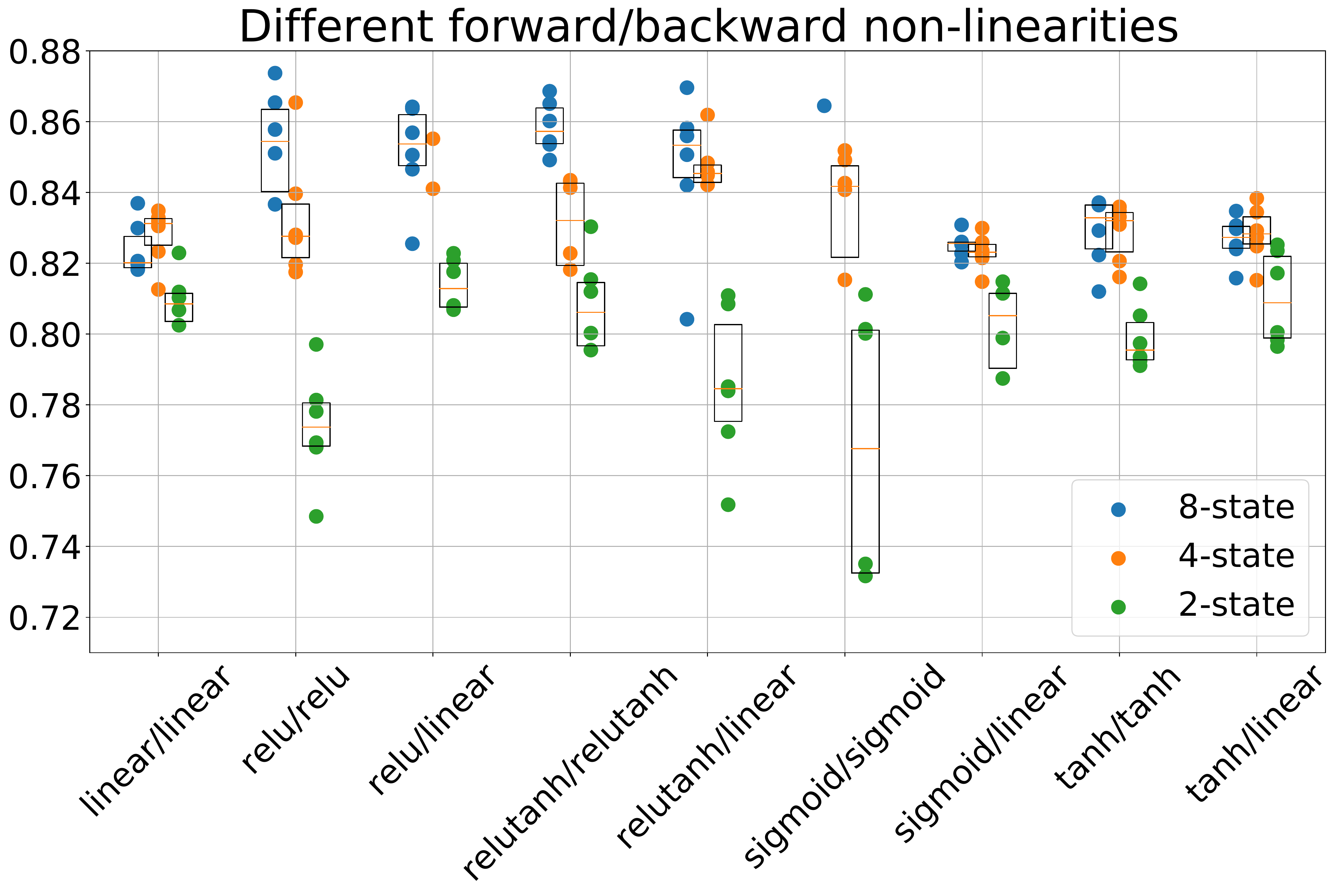}
    \vspace{-2ex}
    \caption{Performance of different non-linearities. All models were meta-trained with two hidden layers and
    use 2-8 states per neuron.}
    \label{fig:non-linearities}
\end{figure}

\subsection{Non-linearities}
On Fig.~\ref{fig:non-linearities} we show the performance of our meta-trained while using different non-linearites. We explore using both identical non-linearities on forward/backward pass as well as not-using
non-linearity in backward-ass at all at all. It can be seen that we successfully meta-learned a configuration
for almost every combination of non-linearities. Also perhaps not-surprisingly fully linear network while correctly does not see any advantage of increasing number of states, while networks with non-linearity improve
with the number of states. 

\begin{figure}
    \centering
    \includegraphics[width=0.46\textwidth]{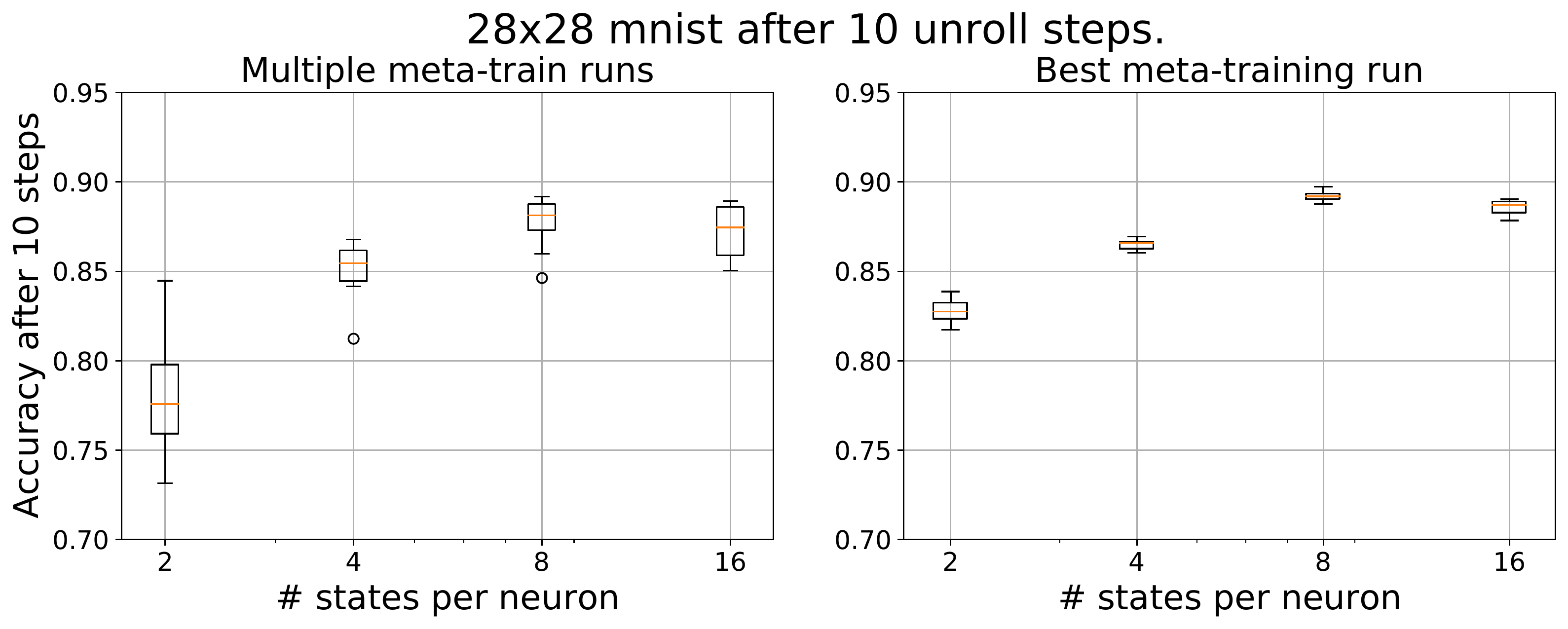}
    \caption{Performance after 10 unrolls with different number of channels per neuron. All meta-trained runs were trained with the same architecture.}
    \label{fig:num_states_trade_off}
\vspace{-2ex}
\end{figure}
\begin{figure}
    \centering
    \includegraphics[width=0.35\textwidth]{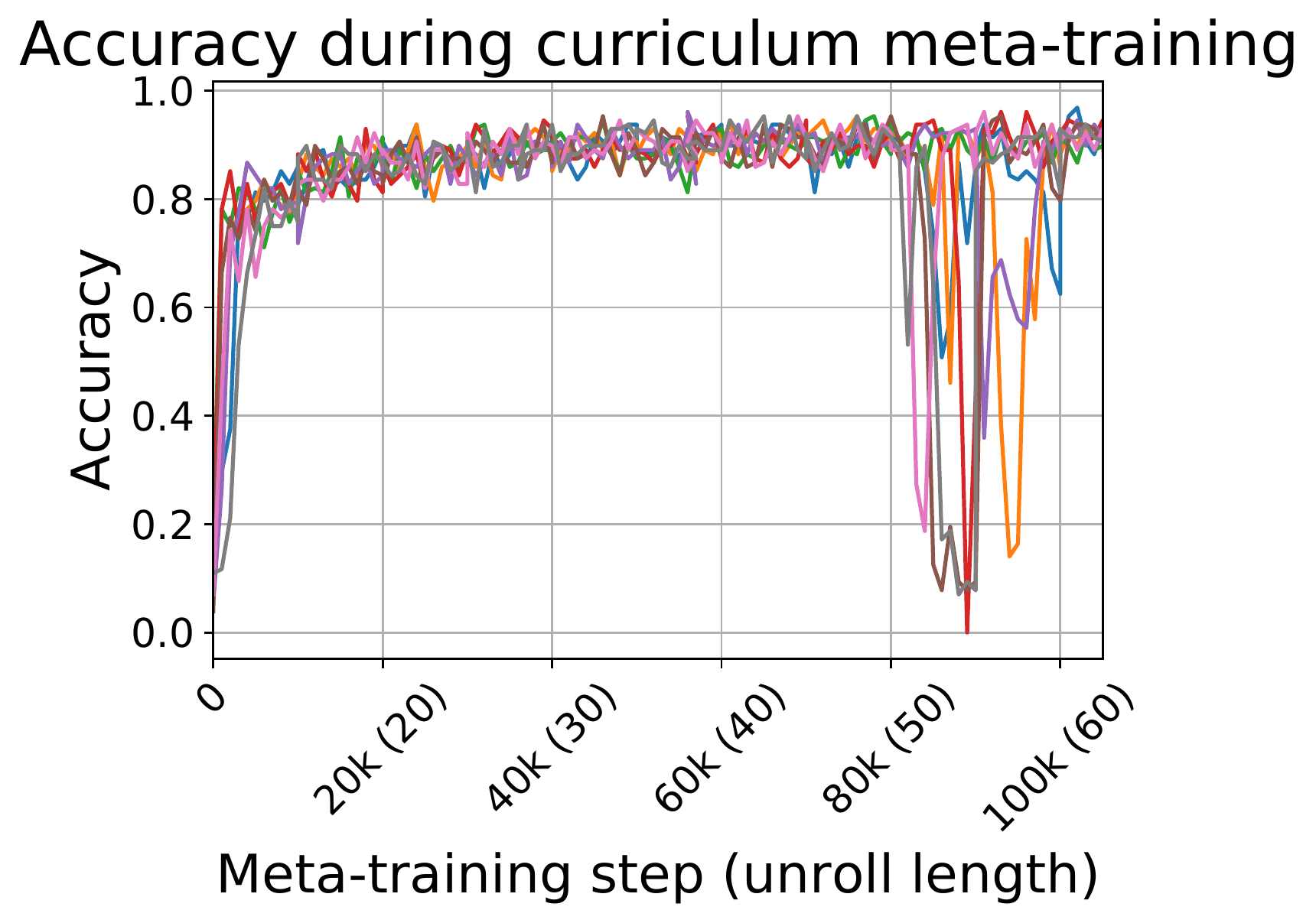}
\vspace{-2ex}        
    \caption{Trajectory of training evolution during curriculum training. }
    \label{fig:mnist-meta-training}
\end{figure}

\subsection{Channels per neuron}
As we mentioned using only 2-states per neuron provides a slightly richer space than gradient descent. So what happens if we increase the number of states? Fig.~\ref{fig:num_states_trade_off} suggests that merely increasing the number of states improves performance, but fully capturing the power of multi-state neurons is subject of future work.  

\subsection{Curriculum metatraining}
Fig.~\ref{fig:mnist-meta-training} shows the accuracy of 8-identical randomly initialized genomes trained for 10,000 steps with initial unroll length of 10 steps. We then increase the length of unroll by 5 steps for each consecutive 10,000 steps and synchronize genomes across all runs. 

\begin{figure}[t]
    \centering
    \includegraphics[width=0.48\textwidth]{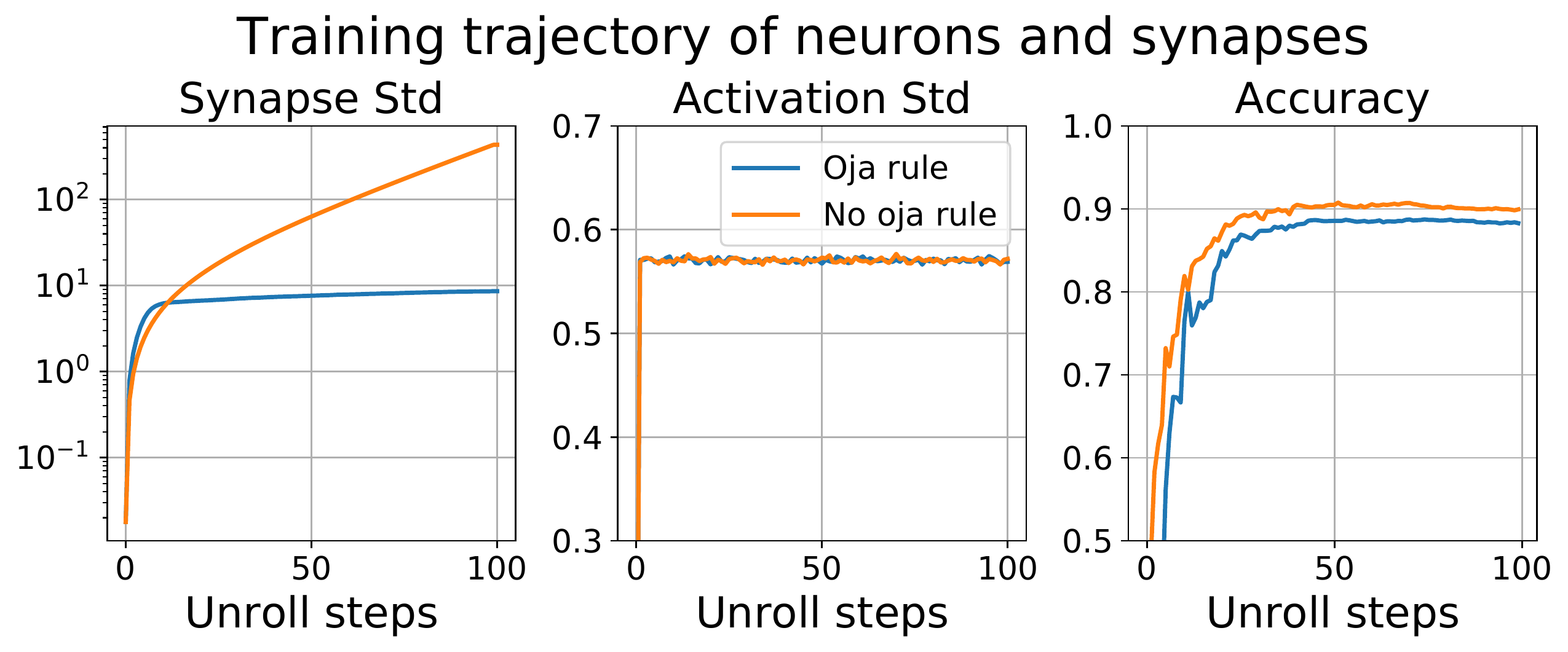}
\vspace{-2ex}
    \caption{The impact of Oja's rule on synapse amplitude.}
    \label{fig:oja_impact}
\end{figure}

\subsection{Importance of Oja Rule} 
In our experiments using additional regularization Oja's term on synapse weights helps preventing synapse explosion.
For instance in Fig.~\ref{fig:oja_impact} we show, that without modified Oja's rule the synapse weights tended to diverge, even though the overall accuracy was not affected until overflow occurred. 

\subsection{Synapse normalization}

\begin{figure}[t]
    \centering
    \includegraphics[width=0.45\textwidth]{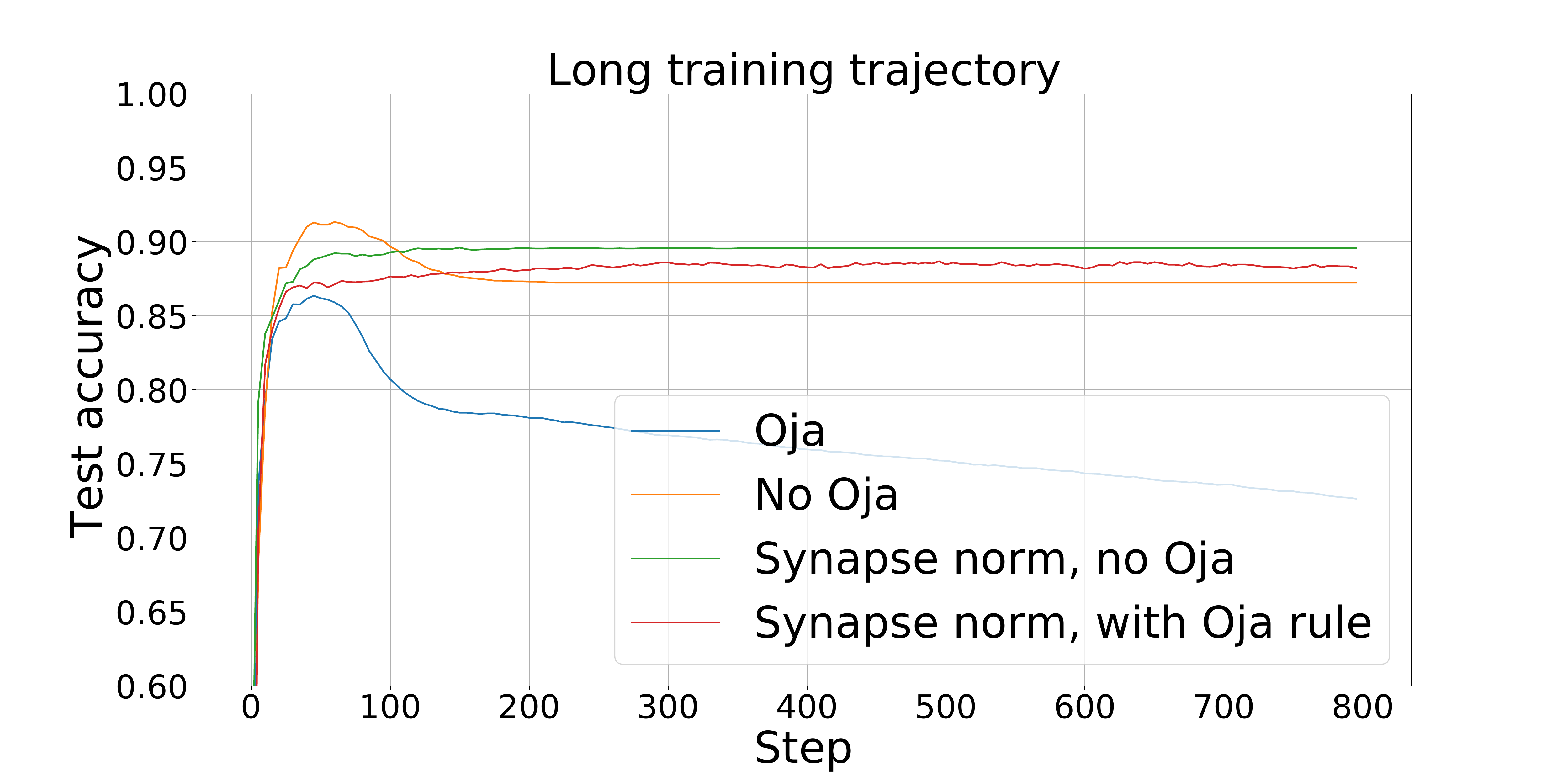}
    \caption{
        \textsc{MNIST} test error measured in the process of training models with and without additional synapse normalization.
        Synapse normalization (``Synapse norm'' label) was only performed when computing neuron activations and did not affect the actual stored synapse values.
        Given stored synapse values $w_{ij}$ with $i$ being the input dimension and $j$ being the output dimension, the effective weight used while computing the activations was chosen as $w'_{ij} = w_{ij} / (\sum_{i'} w_{i'j}^2)^{1/2}$.
    }
    \label{fig:norm-example}
\end{figure}

While reaching a high accuracy at unroll step $n_{\rm target}$ used as a target for meta-learning, the model performance frequently degrades when it is trained beyond that point.
One technique that proved useful for solving this issue is to normalize synapse values during the computation of activations of individual layers.
Fig.~\ref{fig:norm-example} shows evolution of the test accuracy for the best out of 10 runs in 4 different experiments: with and without synapse normalization (and with or without using Oja's rule).
Notice that while the peak accuracy in experiments with normalization may be lower than in those without normalization, it continues to grow well beyond $n_{\rm target}$.
Furthermore, in models with Oja's rule, the synapses saturate during training and if we randomize the labels after step $500$, the model performance degrades soon after, in other words, the model with synapse normalization and Oja's rule continue to learn.
In contrast, synapses grow exponentially in models without the Oja's rule, which prevents them from learning after a certain stage and later their synapses overflow and they eventually diverge.

\subsection{Ablation study}
\begin{figure*}[t]
    \centering
    \begin{tabular}[c]{@{\hspace{0.0\linewidth}}c|@{\hspace{0.00\linewidth}}c@{\hspace{0.01\linewidth}}||@{\hspace{0.01\linewidth}}c|@{\hspace{0.0\linewidth}}c@{\hspace{0.0\linewidth}}}
    \multicolumn{2}{c}{Meta-Train: MNIST} & \multicolumn{2}{c}{Meta-Train: 10-way Omniglot} \\\hline
    Meta-Eval: MNIST & Meta-Eval: Omniglot & Meta-Eval: MNIST & Meta-Eval: Omniglot \\\hline\hline
    \multicolumn{4}{c}{Backward update: Additive} \\
    \includegraphics[width=0.23\textwidth]{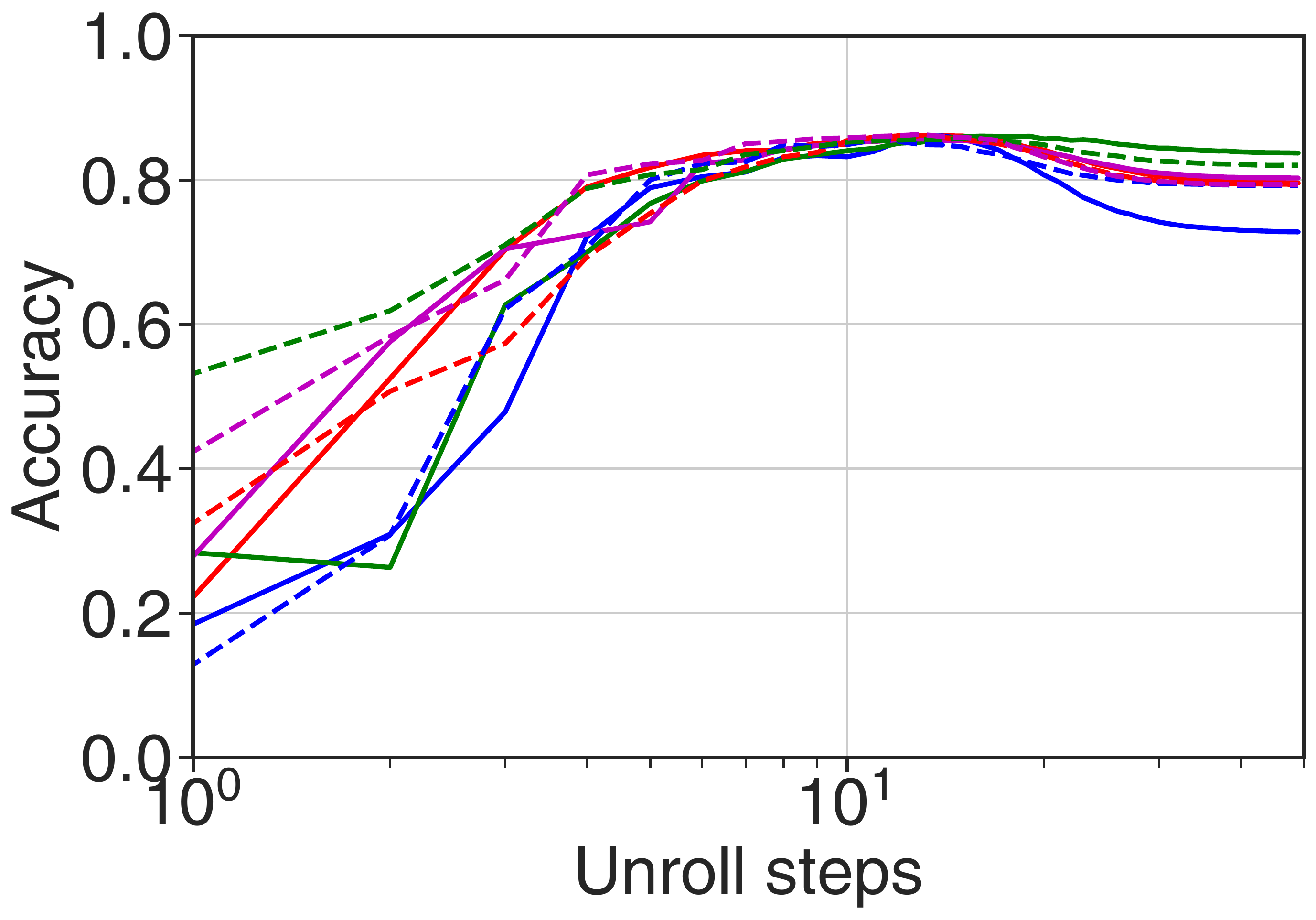} &
    \includegraphics[width=0.23\textwidth]{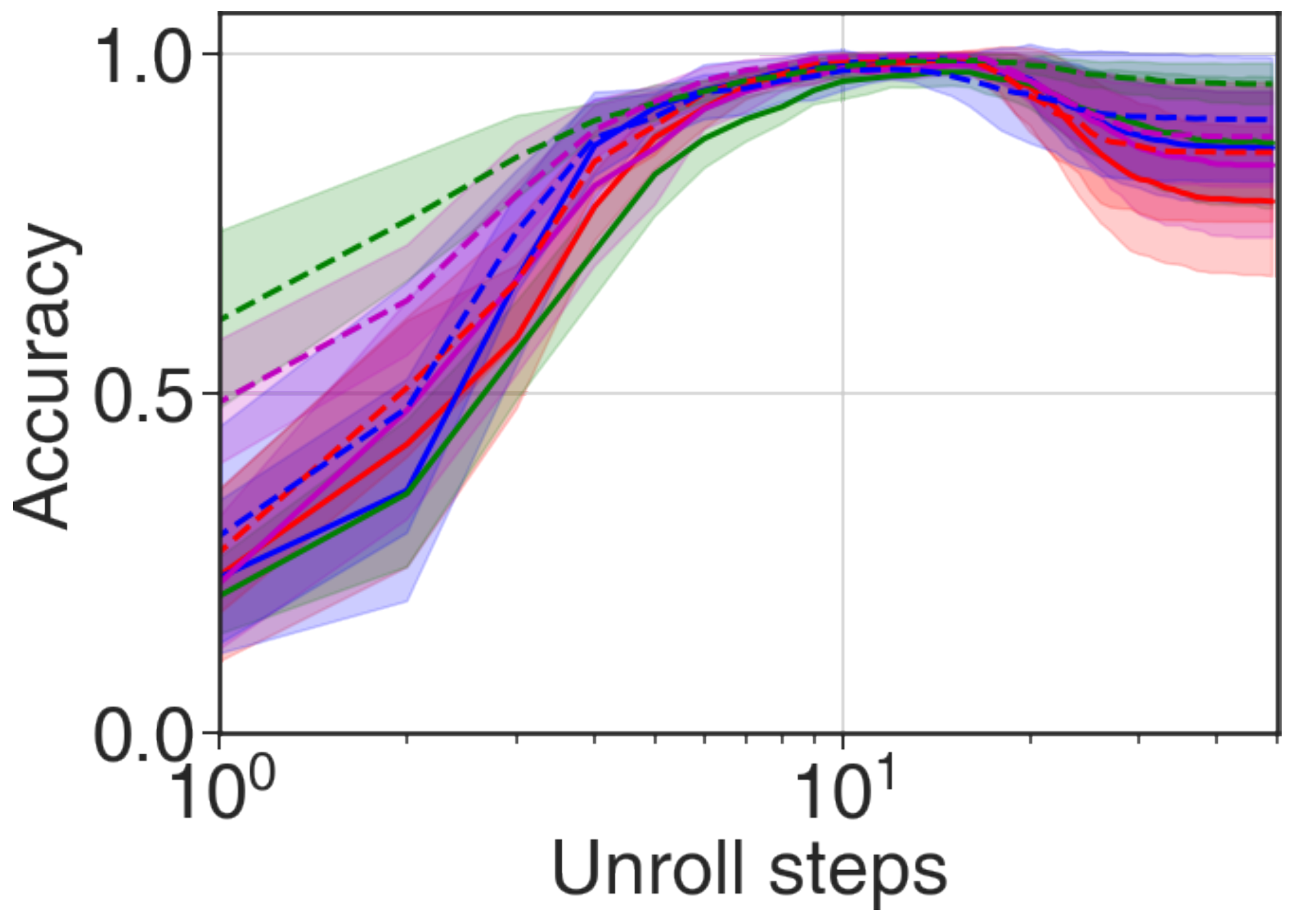}
     &
    \includegraphics[width=0.23\textwidth]{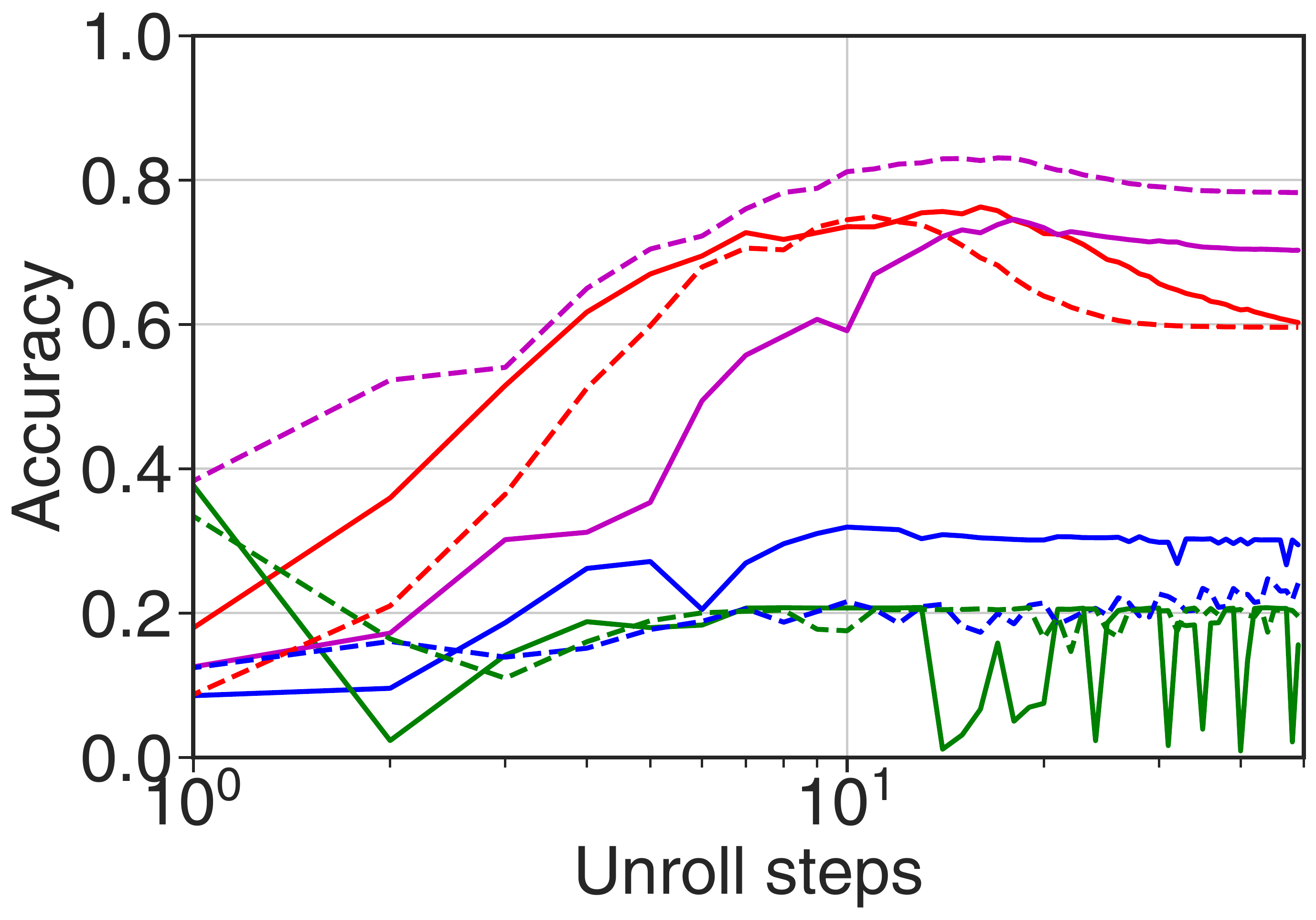} & 
    \includegraphics[width=0.23\textwidth]{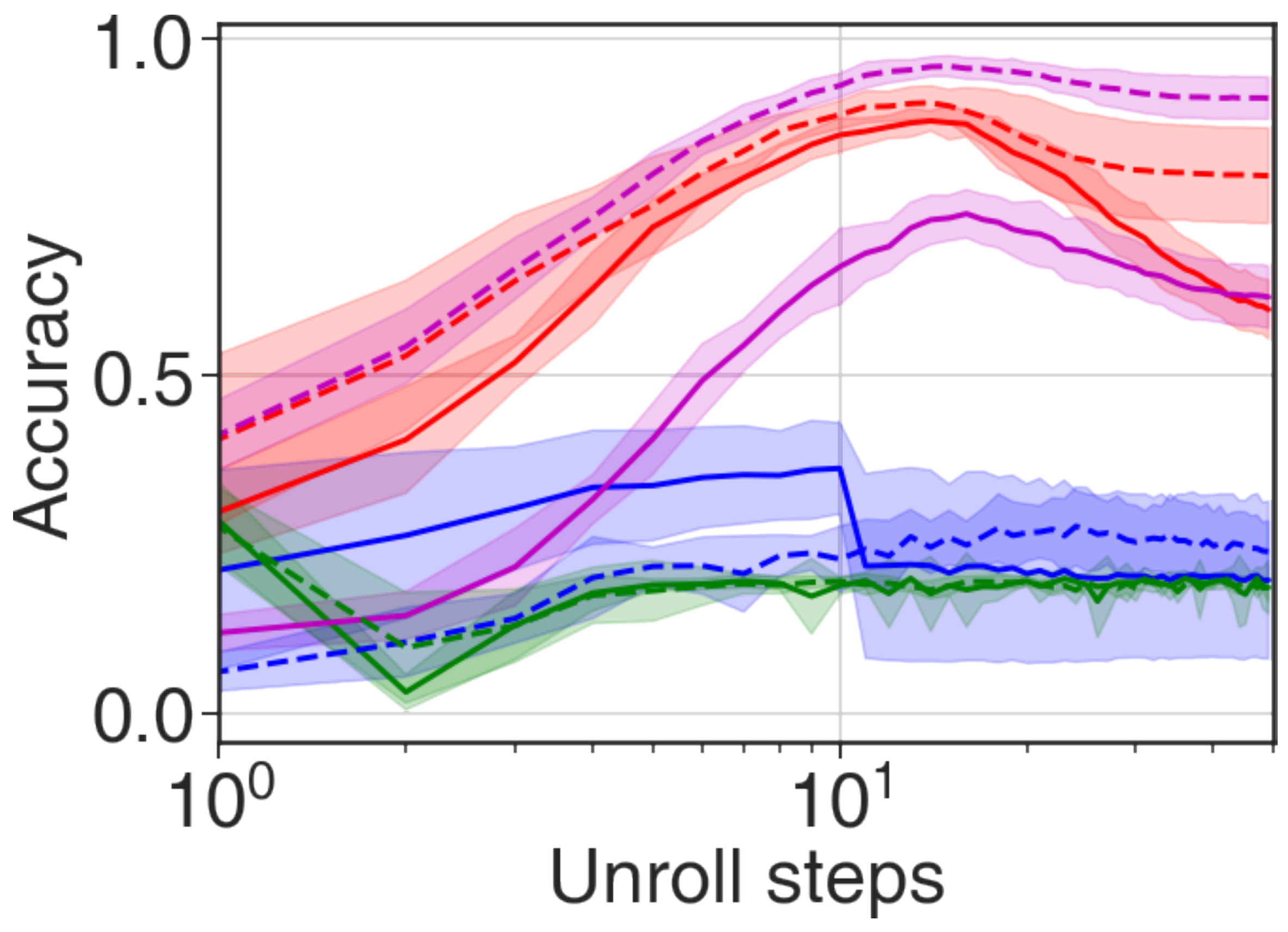} \\\hline
    \multicolumn{4}{c}{Backward update: Multiplicative} \\
    \includegraphics[width=0.23\textwidth]{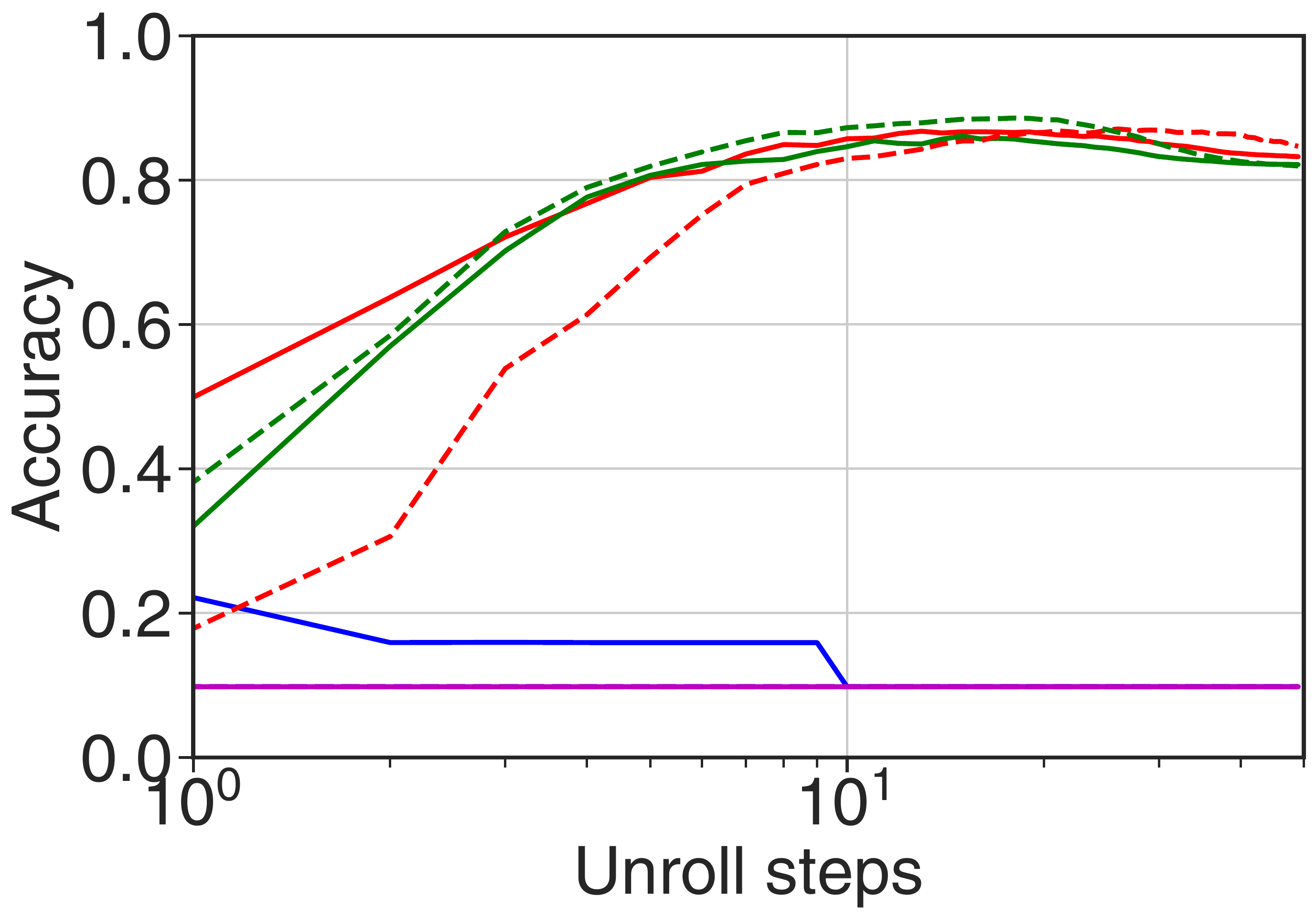} &
    \includegraphics[width=0.23\textwidth]{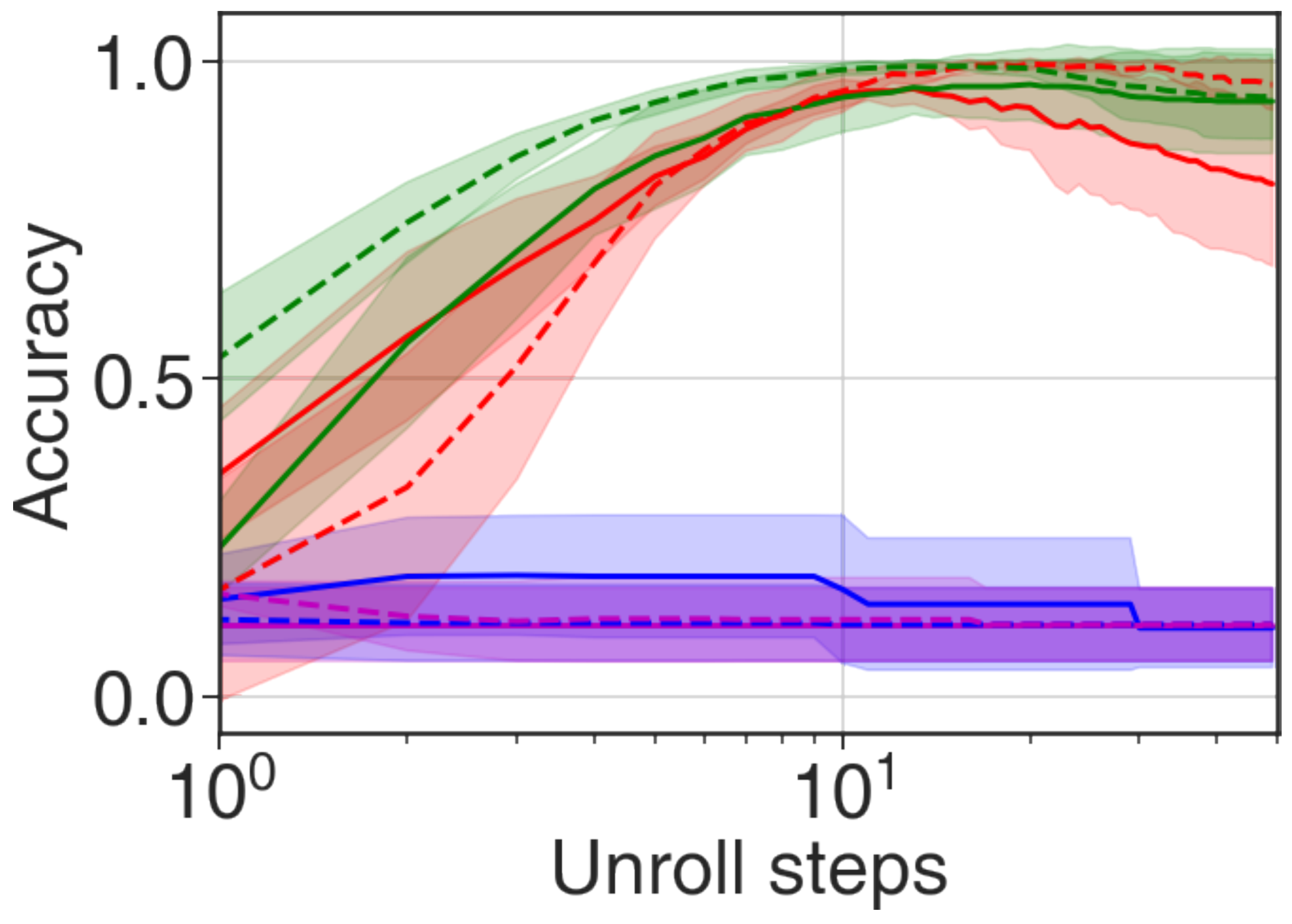} &
    \includegraphics[width=0.23\textwidth]{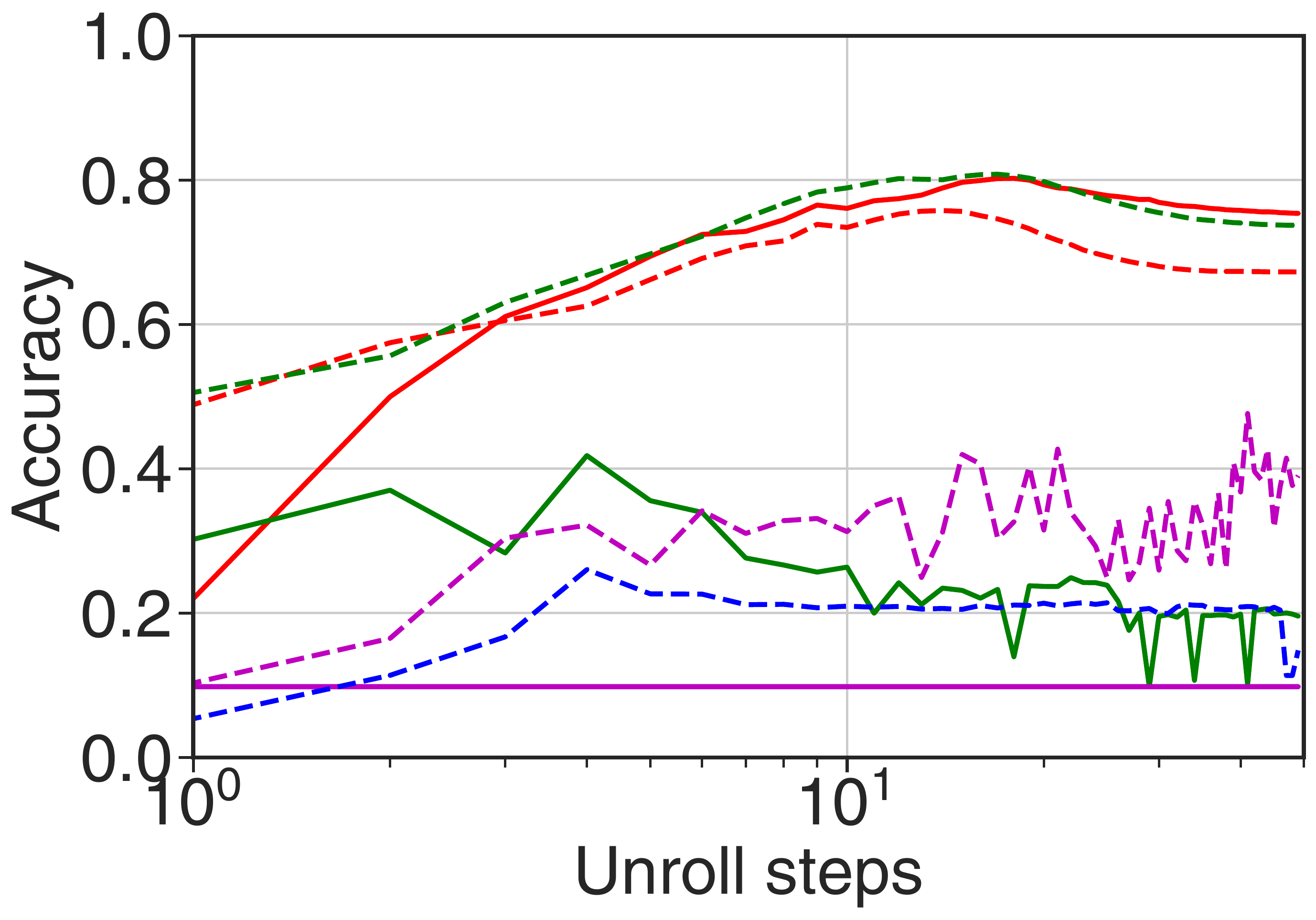} & 
    \includegraphics[width=0.23\textwidth]{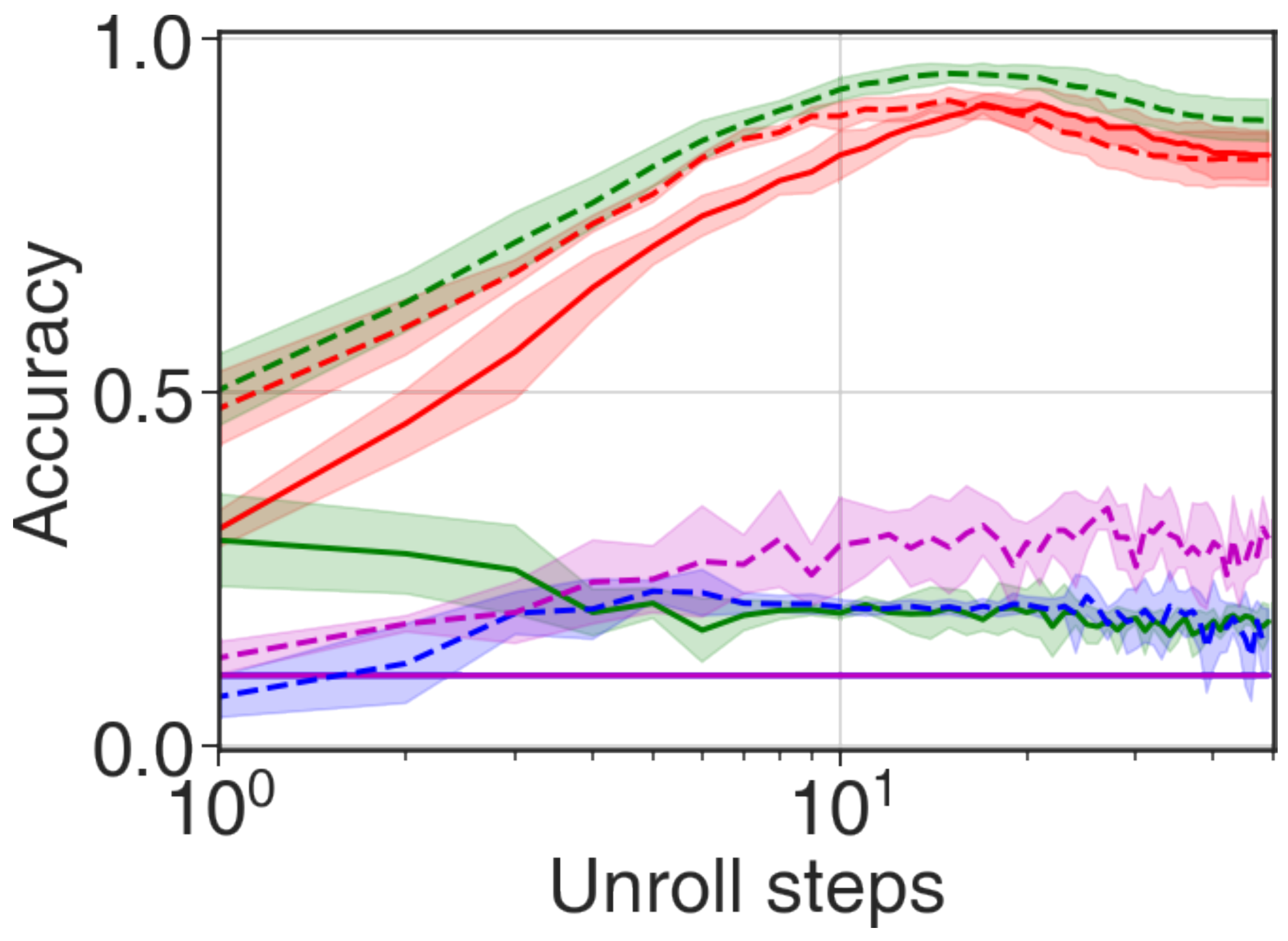} \\\hline
    \multicolumn{4}{c}{Backward update: Multiplicative second state only} \\
    \includegraphics[width=0.23\textwidth]{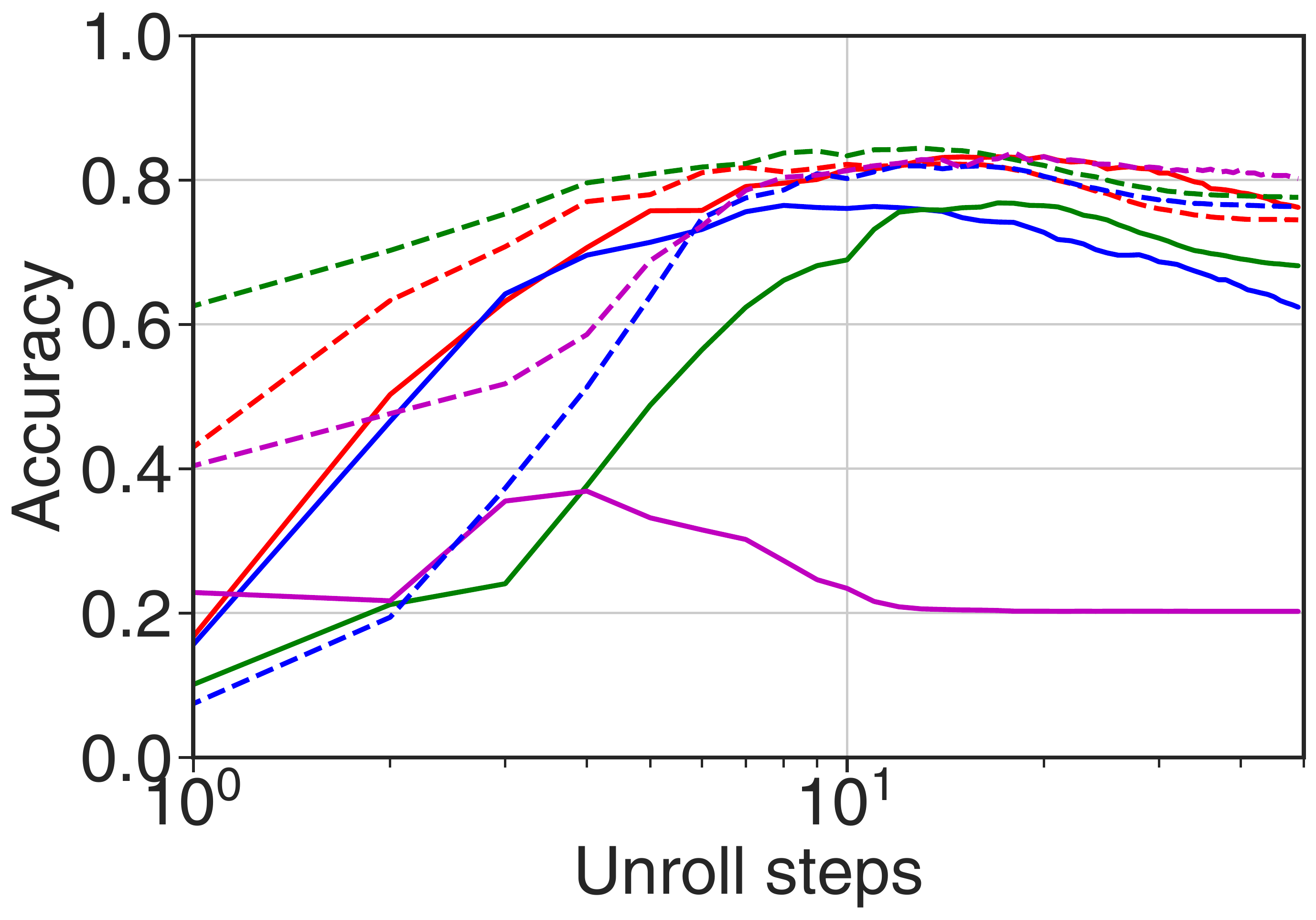} &
    \includegraphics[width=0.23\textwidth]{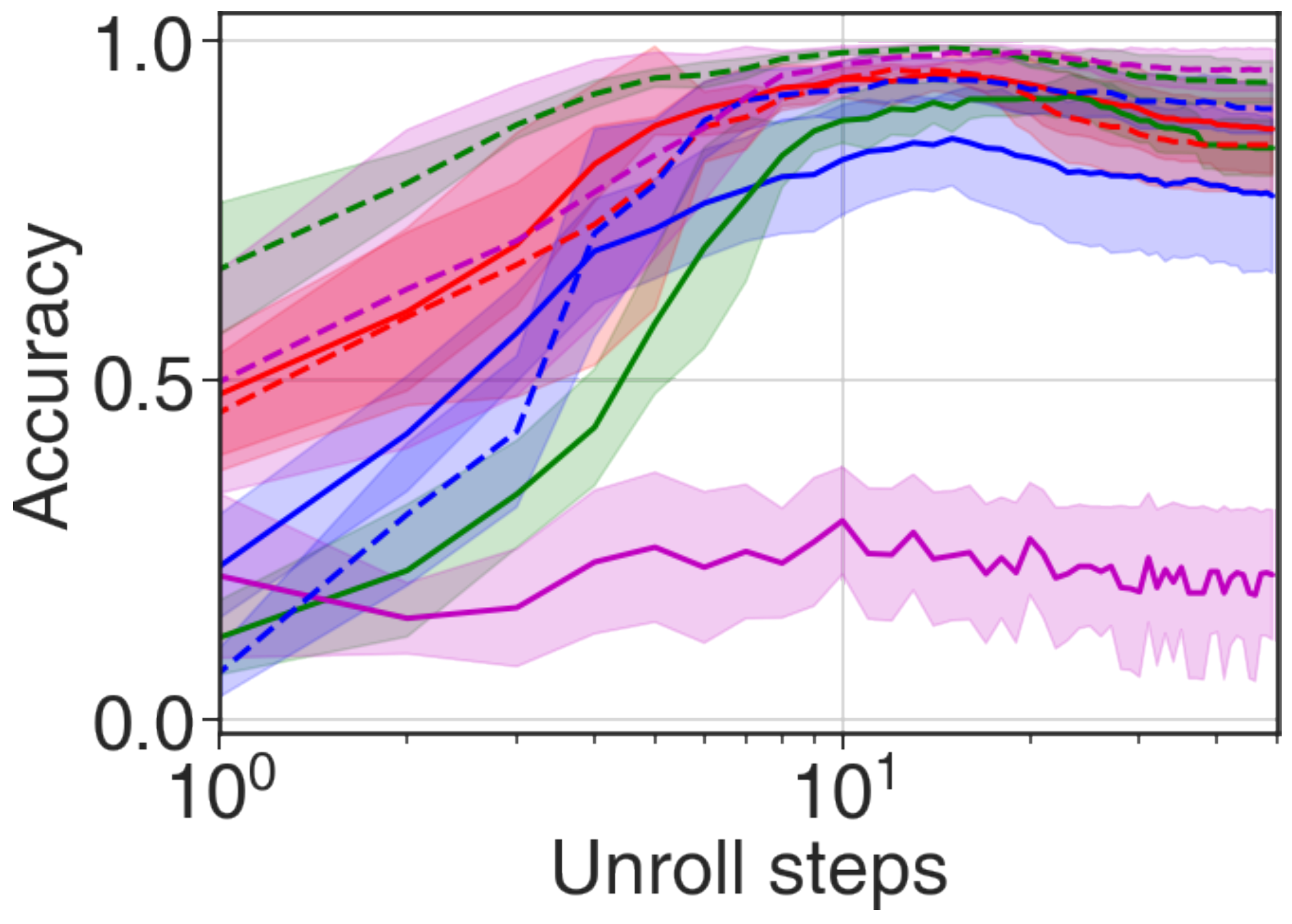}&
    \includegraphics[width=0.23\textwidth]{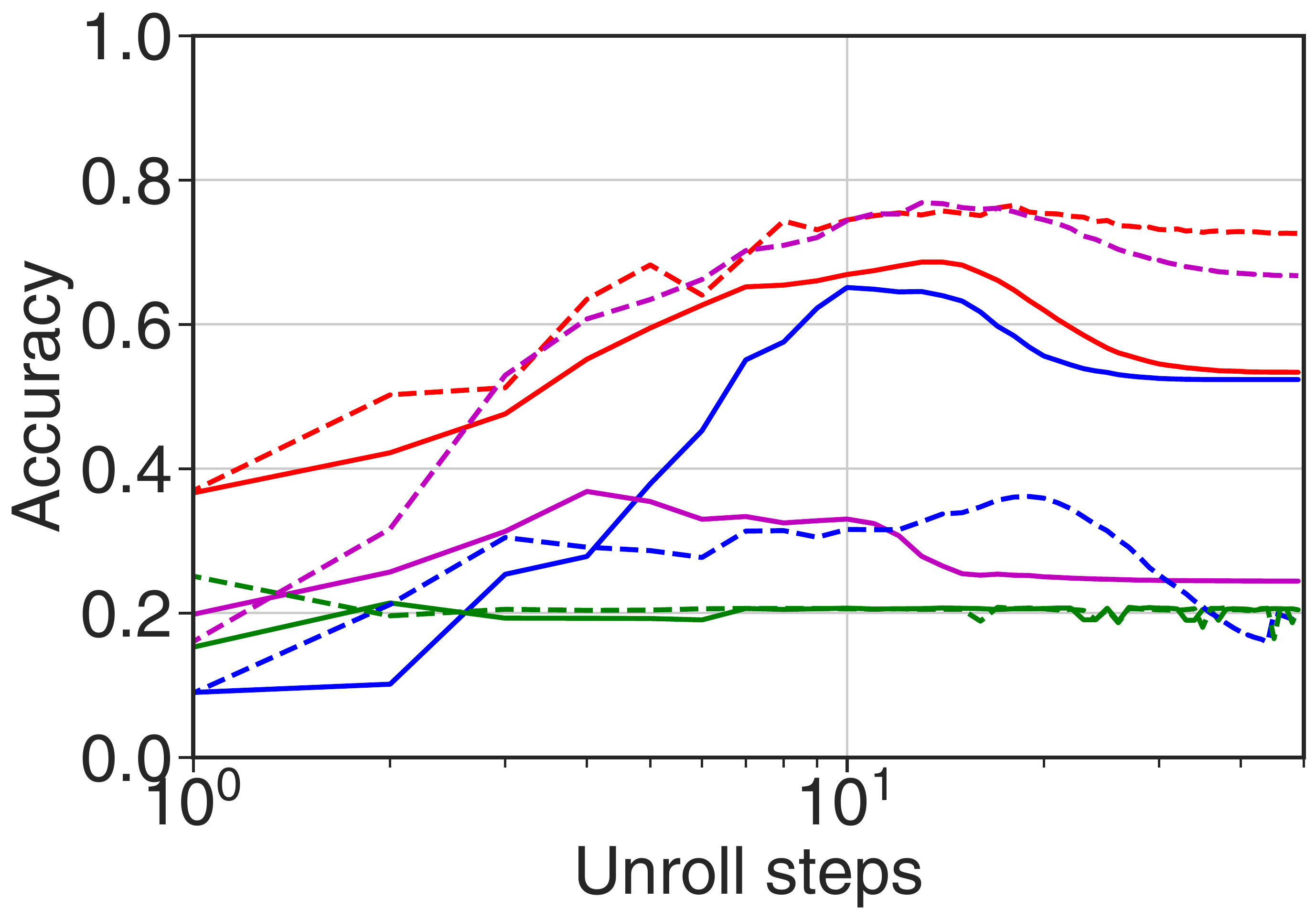} & 
    \includegraphics[width=0.23\textwidth]{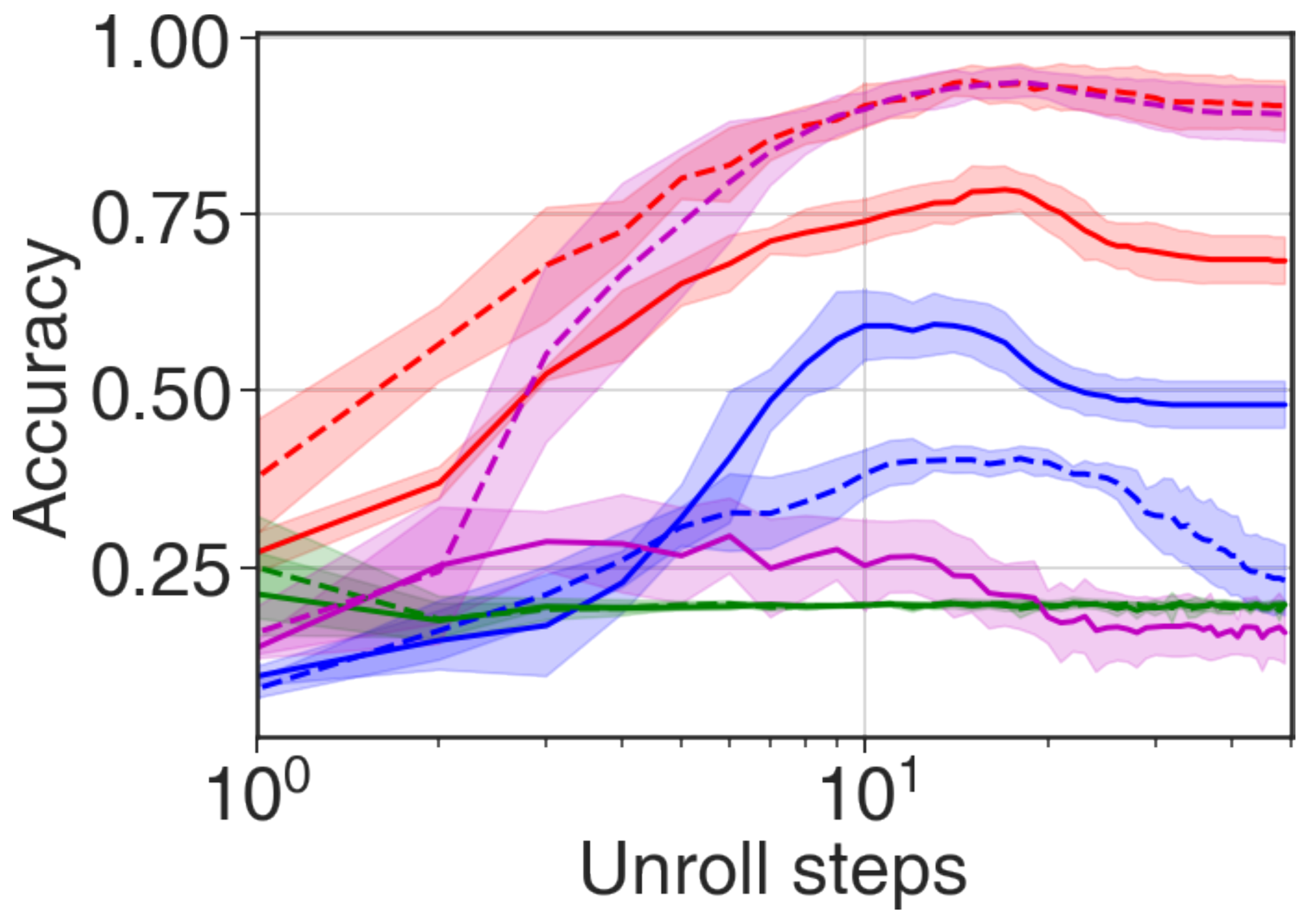} \\        
    \end{tabular}
    \includegraphics[width=0.8\textwidth]{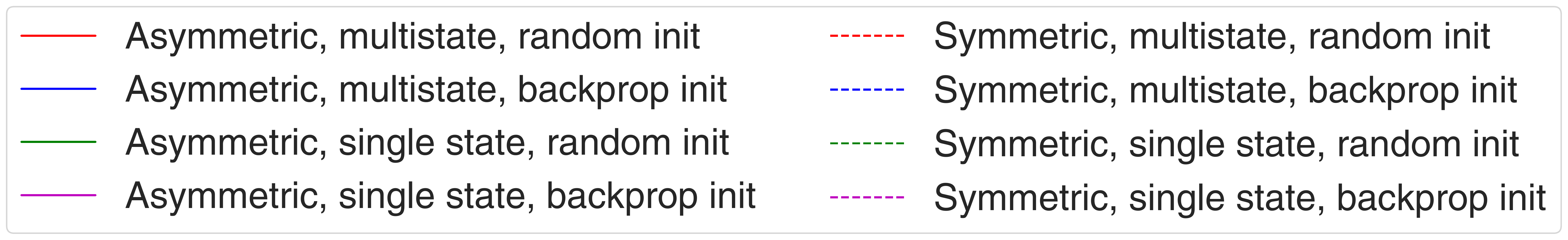} \\        
    \caption{Testing different variants of our proposed learning method. See text for the description of each parameter. All variants were trained on MNIST or Omniglot to 10 unrolls and evaluated on MNIST and 10 classes from Omniglot. Errorbars show standard deviation of the accuracy over 10 different subsets of Omniglot. Only the best result of 8 runs is plotted.}
    \label{fig:ablation_symmetry}
\end{figure*}

In Fig.~\ref{fig:ablation_symmetry} we show the ablation study for the following parameters:
\begin{itemize}
    \item Type of the backward update: \emph{additive}, \emph{multiplicative} or \emph{multiplicative second state only}. This refers to the states' update on the backward pass.
    \item Forward and backward synapses: \emph{symmetric} vs \emph{asymmetric}.
    \item Number of states in synapses: \emph{single state} vs \emph{multistate}. 
    \item Initialization: \emph{random} vs \emph{backprop genome initialization}.
\end{itemize}

The backpropagation algorithm can be recovered as an initialization to \emph{``Multiplicative second state only, symmetric, single state, backprop init''} variant (dashed purple line). The most general version of the algorithm that is used in most experiments in the paper is  \emph{``Additive, symmetric, multistate, random init''} variant (solid red line).

Depending on which backward update rule is chosen, different combinations of parameters dominate over others. While we couldn't find a clear pattern to order the parameters by their performance, surprisingly, most of the variants give reasonable results, suggesting that the space of potentially useful update rules is quite large.

\end{document}